\pdfoutput=1

\documentclass[11pt]{article}

\usepackage[final]{acl}

\usepackage{times}
\usepackage{latexsym}

\usepackage[T1]{fontenc}

\usepackage[utf8]{inputenc}

\usepackage{microtype}

\usepackage{inconsolata}

\usepackage{graphicx}

\usepackage{booktabs}
\usepackage{subfig}
\usepackage{adjustbox}
\usepackage{tikz}
\def\checkmark{\tikz\fill[scale=0.4](0,.35) -- (.25,0) -- (1,.7) -- (.25,.15) -- cycle;}
\title{Not a nuisance but a useful heuristic:\\Outlier dimensions favor frequent tokens in language models}

\author{
 \textbf{Iuri Macocco\textsuperscript{1}},
 \textbf{Nora Graichen\textsuperscript{1}},
 \textbf{Gemma Boleda\textsuperscript{1,2}},
 \textbf{Marco Baroni\textsuperscript{1,2}}
\\
 \textsuperscript{1}Universitat Pompeu Fabra,
 \textsuperscript{2}ICREA
\\
 \small{
   \texttt{\{firstname.lastname\}@upf.edu}
 }
}

\begin{document}
\maketitle
\begin{abstract}
We study last-layer \textit{outlier dimensions}, i.e.\ dimensions that display extreme activations for the majority of inputs. We show that outlier dimensions arise in many different modern language models, and trace their function back to the heuristic of constantly predicting frequent words. We further show how a model can block this heuristic when it is not contextually appropriate, by assigning a counterbalancing weight mass to the remaining dimensions, and we investigate which model parameters boost outlier dimensions and when they arise during training. We conclude that outlier dimensions are a specialized mechanism discovered by many distinct models to implement a useful token prediction heuristic.
\end{abstract}

\section{Introduction}

\begin{figure*}[ht]
    \centering
    \includegraphics[scale=0.45]{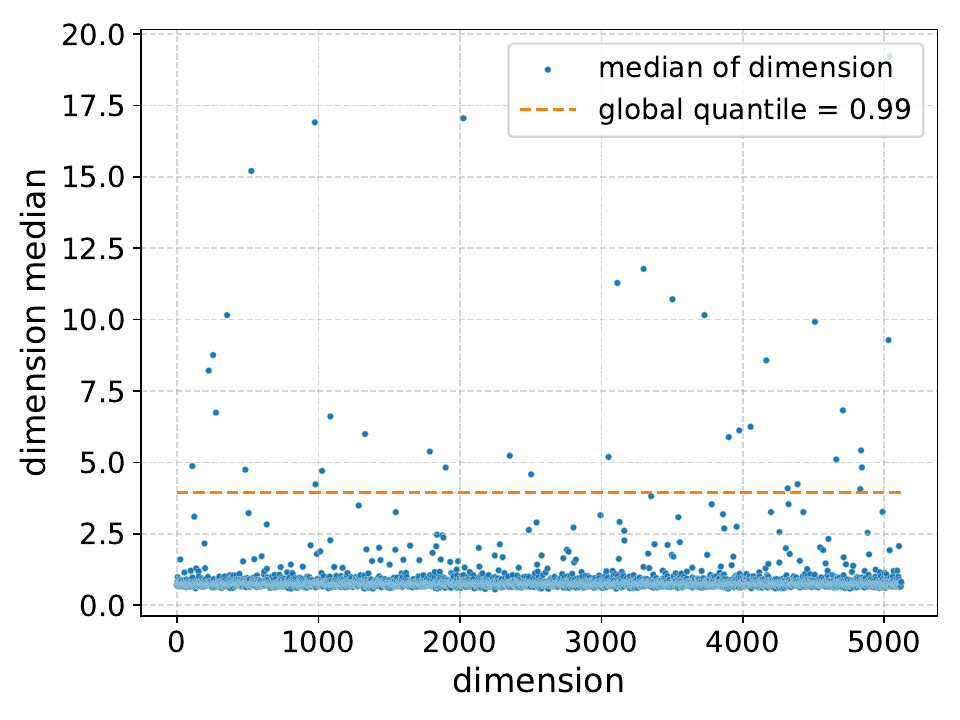}\hfill
    \includegraphics[scale=0.45]{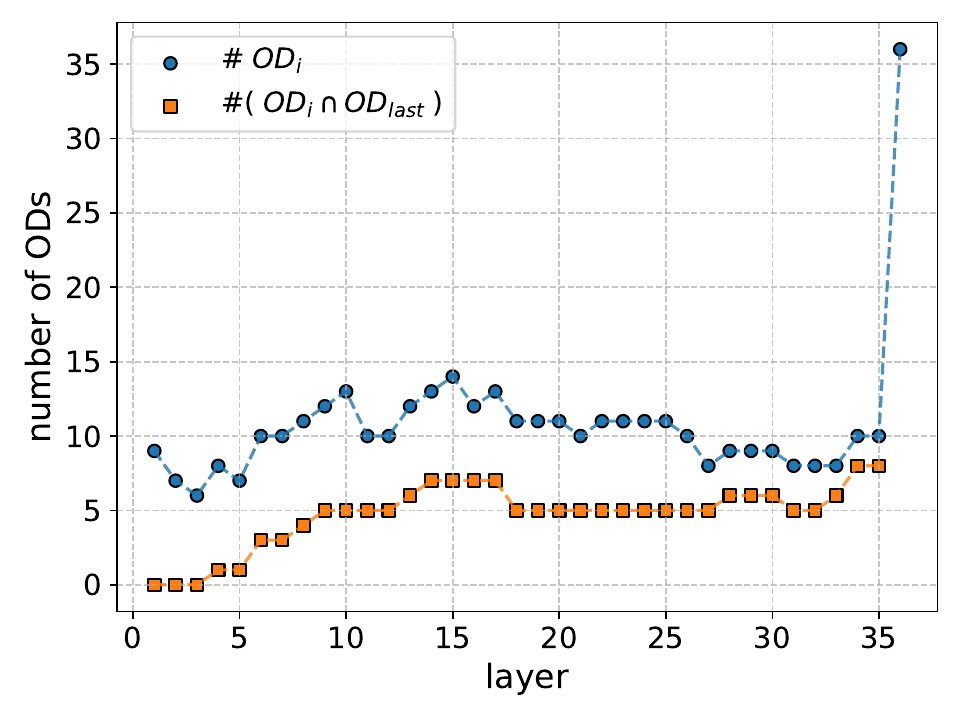}
    \caption{\small \textbf{Left:} Median activation values across our dataset (see Sec.~\ref{sec:methodology}) for each last-layer dimension of pythia-12b. The orange line separates the top 1\% of values across all dimensions, used to assess whether a dimension is an outlier. \textbf{Right:} Evolution of outliers across the layers. The blue dots count the total number of outlier dimensions (ODs) per layer; the orange squares represent the number of outliers that are also ODs in the last layer (omitted in the last layer because they are the same by definition).}
    \label{fig:ods}
\end{figure*}

It has been widely reported that modern language models (LMs) present a number of extreme-distribution phenomena, with some parameters and activations that are systematically much larger than the others \citep[e.g.,][]{kovaleva-etal-2021-bert,timkey-van-schijndel-2021-bark}. In this paper, we investigate one of these phenomena, namely the presence of \textit{outlier dimensions} (ODs) on the last layer of LMs. Unlike what \citet{sun2024massiveactivationslargelanguage} called \textit{massive activations}, that only occur for specific input tokens, ODs are dimensions that display very extreme activation values for a majority of the inputs, as shown for the pythia-12b model in the left panel of Fig.~\ref{fig:ods}. While dimensions with similar properties also occur in earlier layers, we focus on those in the last layer because ODs tend to be more common in the last layer and, importantly, most of the ODs in the last layer are not outliers in earlier layers, which suggests that they play a role specific to output generation (see ODs by layer in Fig.~\ref{fig:ods}-right for pythia-12b, and Fig.~\ref{fig:ods_overlap_all} in Appendix \ref{app:ods-layers} for the other models we experiment with). 

This hypothesis is borne out in the results. Across a number of LMs, we find that ablating ODs significantly affects model performance and, specifically, that they are part of an \textit{ad-hoc} mechanism boosting the prediction of frequent tokens---a sensible heuristic given the skewed nature of word frequency distributions \citep{Baayen:2001}. Indeed, keeping \textit{only} the ODs makes the LMs predict just a few very common tokens, a strategy that results in very low but non-negligible accuracy. Moreover, we show how ODs interact with the unembedding matrix to favor frequent tokens in general, but also how the cumulative effect of the other dimensions can outweigh the effect of ODs when a non-OD-favored token must be predicted. We also present evidence that OD values are pushed up by the main directions in the space of the last-layer MLP down-projection matrix, as well as by high biases and weights in the last layer normalization. Finally, we show how the ODs emerge early during training after a first phase in which the model is predicting frequent words by other means.

Our main contributions are as follows: 
\begin{itemize}
    \item We present a thorough characterization of ODs in a set of modern LMs.
    \item We identify the \textit{function} of these activations, showing that they generally work as a specialized module for frequent-token prediction.
    \item We describe the \textit{mechanics} by which frequent-token prediction is achieved (or blocked) through the interaction between ODs and the unembedding matrix.
\end{itemize}

The code to reproduce all the results is available at \url{https://github.com/imacocco/llms_outlier_dimensions/}

\section{Related work}
Extreme values in transformer-based LM activations and weights 
are of interest due to their negative impact on quantization
\citep{dettmers2022,bondarenko2023quantizabletransformersremovingoutliers}, and their role in shaping anisotropic representations, which are linked, for instance, to poorer performance on semantic similarity tasks \citep[e.g.,][]{ethayarajh-2019-contextual, Haemmerl_Fastowski_Libovický_Fraser_2023}. Intriguingly, just a few dimensions with such extreme values challenge 
model compression and general quantization performance, and their removal has been linked to general performance degradation \citep[e.g.,][]{kovaleva-etal-2021-bert,zeng2023glm130bopenbilingualpretrained}.

\citet{puccetti-etal-2022-outlier} find a positive correlation between input token frequency and the activation magnitude of extreme dimensions in hidden layers, which diminishes in later layers.
Notably, they qualitatively observe a tendency for BERT to predict more frequent tokens when extreme dimensions are removed—which is the opposite of our findings (see e.g.\ Fig.~\ref{fig:prediction-corpus-frequency-pythia}).
\citet{puccetti-etal-2022-outlier} use BERT, whereas we focus on decoder-only transformer models, and they employ different criteria to identify ODs. These methodological differences may explain the discrepancies in our results.

\citet{sun2024massiveactivationslargelanguage} and \citet{An_Zhao_Yu_Tang_Wang_2025} identify a specific subset of extreme values, referred to as \textit{massive activations} or \textit{systematic outliers}. These are linked to the phenomenon of ``attention sinking'', a mechanism that reduces the contribution of attention heads in contexts where they are not useful (see also \citet{Cancedda:2024}). Additionally, they function as context-aware scaling factors, modulating the influence of certain tokens \citep{An_Zhao_Yu_Tang_Wang_2025}.

Importantly, this specific subset of extreme values (massive activations/systematic outliers) differs from what we refer to as ODs in key ways: (i) they appear only in specific positions or for specific tokens, such as punctuation marks; (ii) their values are even more extreme (1,000 times the mean); (iii) they are very few (typically four or fewer per model across all layers, in contrast to the number of last-layer ODs we report in  Table \ref{tab:ablation_results}); and (iv) they tend to emerge more prominently in the middle layers, while we focus on the last layer.\footnote{Indeed, applying the criterion of \cite{An_Zhao_Yu_Tang_Wang_2025} to our models, we don't find any massive activation on the last layer, which is the one where our ODs occur.} %
Furthermore, these activations often correspond to punctuation marks or frequent words as input tokens, whereas ODs tend to occur across the board, but might favor items from these classes as \textit{output} tokens.

\section{Methodology}
\label{sec:methodology}

\paragraph{Models and datasets}
\label{par:models}
We experiment with the following pre-trained language models, all available on HuggingFace:\footnote{\url{https://huggingface.co/}} pythia-12b(-deduped) \citep{biderman2023pythia}, mistral-7b \citep{jiang2023mistral7b}, llama3-8b \citep{grattafiori2024llama3herdmodels}, olmo2-13b \citep{olmo20252olmo2furious}, qwen-14b \citep{qwen2.5}, opt-13b \citep{zhang2022optopenpretrainedtransformer}, gemma-9b \citep{gemmateam2024gemma2improvingopen}, and stable-12b \citep{bellagente2024stable}. These were the latest and largest instances of each model family that we could analyze in a reasonable time, given our computing resources. See Table \ref{tab:MBHyper} in 
Appendix \ref{app:language-model-details} for further model details.

We extract our input data from WikiText-103 \citep{merity2016pointersentinelmixturemodels}. Specifically, we sample 50k non-overlapping fragments of 101 (typographic) words which do not cross document boundaries. All fragments start at the beginning of a sentence, whereas their final token is not required to coincide with and end-of-sentence boundary (so, it can belong to any part-of-speech). The first 100 words are fed to the models as contexts, and the first token of the last word constitutes the ground truth against which to compare model predictions.

\paragraph{OD identification} We define outlier dimensions (ODs) as dimensions that have extreme values across a large variety of inputs. We operationalize this as follows. We consider extreme values those that are in the top 1\% when ranking the absolute values of the activations of all dimensions in a given layer for our whole dataset (i.e., the ranked list will be of length $n\times50$k, where $n$ is the number of dimensions in the relevant layer of the model of interest). 
We then define an OD as a dimension whose median activation across the 50k samples in our dataset is an extreme value. That is, an OD dimension will display an activation that is among the 1\% more extreme across all inputs and dimensions for at least 50\% of the inputs. Table \ref{tab:ablation_results} (leftmost column) reports the number of ODs we find with this method in the last layer of each model (see Fig.~\ref{fig:ods} above and Fig.~\ref{fig:ods_overlap_all} in Appendix \ref{app:ods-layers} for the other layers).
The \textit{z-score} column of the same table shows that the ODs identified in this way not only have values that ranked among the most extremes, but are also high in absolute terms—at least eight standard deviations above the mean on average for all models. Note also that ODs are a very small proportion of the dimensions in a given model (between 4 and 36, whereas layers are of dimensionality 3584-5120 depending on the model, cf.\ Appendix~\ref{app:language-model-details}).

\section{Results} 

\begin{table*}[htb]
    \centering
    \small
    \setlength{\tabcolsep}{4pt} 
    \begin{tabular}{l|cc|ccc|cccc}
        \toprule
        & \multicolumn{2}{c}{ODs} & \multicolumn{3}{c}{Accuracy [\%]} & \multicolumn{4}{c}{\# of Distinct predicted Tokens}  \\
        \midrule
        Model & \# & z-score & FM & abl-OD & only-OD & FM & abl-OD & only-OD & only-rnd\\
        \midrule
        pythia-12b & 36 & 9.8 ± 5.2 & 43.0 & 34.3 & 4.7 & 7504 & 11116 & 195 & 7254 ± 269 \\
        mistral-7b & 28 & 9.6 ± 4.6 & 46.6 & 41.9 & 1.7 & 6334 & 7098 & 712 & 4518 ± 427 \\
        llama-8b   & 12 & 11.9 ± 5.7 & 49.2 & 41.8 & 1.2 & 8034 & 12030 & 150 & 2367 ± 339 \\
        olmo2-13b  & 24 & 10.8 ± 7.6 & 52.1 & 41.0 & 7.4 & 8317 & 13320 & 655 & 5787 ± 1062 \\
        qwen-14b   & 38 & 8.8 ± 7.1 & 49.9 & 32.2 & 1.3 & 8171 & 11233 & 710 & 10236 ± 750 \\
        opt-13b    &  4 & 29.1 ± 11 & 42.7 & 42.5 & 0.0 & 7911 & 9026 & 85 & 281 ± 13 \\
        gemma-9b   &  6 & 17.2 ± 10.7 & 48.4 & 48.0 & 0.1 & 8229 & 8539 & 79 & 373 ± 46 \\
        stable-12b & 23 & 9.2 ± 6.8 & 49.4 & 30.2 & 0.0 & 8029 & 1821 & 96 & 947 ± 109 \\
        \bottomrule
    \end{tabular}
    \caption{\small \textbf{ODs:} Number of ODs and their average z-score with respect to the mean of absolute values of all last-layer activations.
    Effects of OD/non-OD ablation. \textbf{FM:} full model; \textbf{abl(ate)-X:} dimensions X were set to 0; \textbf{only-X:} dimensions other than X were set to 0. Random-dimension experiments are repeated with 10 different seeds and averaged. 
    Ablate-random experiments, where the same number of random dimensions were ablated as there are ODs, always resulted in values matching those of the full model up to the third significant digit for both accuracy and \# of distinct tokens, and we do not report them in the table. Only-random (only-rnd) experiments keep the same number of random dimensions as there are ODs; their accuracies are not reported either, as they are typically $\sim 0.01\%$ and always $<0.1\%$.}
    \label{tab:ablation_results}
\end{table*}

The remaining columns of Table~\ref{tab:ablation_results} suggest that there is a group of models with the same pattern of behavior across the board (pythia-12b, mistral-7b, llama-8b, olmo2-13b, qwen-14b) and three models that often deviate in different ways (opt-13b, gemma-9b, stable-12b).

\paragraph{Impact of outlier dimensions on predictions} ODs greatly impact model prediction: Ablating the outlier dimensions, i.e.\ setting them to 0, leads to a large decrease in accuracy for most models, despite the fact that the ODs are a very small proportion of the total dimensions of the models (see column \textit{abl-OD}). The exceptions are opt-13b and gemma-9b, which experience very small decreases. Instead, ablating a comparable number of random dimensions in the last layer never affects accuracy by more than 0.1\% for any model (results not shown in the table).

The symmetric operation, namely drastically ablating the model by setting all last-layer dimensions but ODs to 0, causes performance to radically drop, as could be expected (see column \textit{only-OD}). However, in most cases the performance is still much higher than in the only-random condition, where a comparable number of randomly sampled dimensions is ablated (this yields accuracies $<0.1\%$, not shown in the table). Exceptions are opt-13b, gemma-9b, and stable-12b, with accuracies near 0 also in the only-OD condition.
\footnote{Table \ref{tab:surprisal} in Appendix \ref{app:surprisal} reports the ablation effects in terms of surprisal, which exhibits the same trends.}

\paragraph{By which means does ablating ODs affect accuracy in most models?} The rightmost panel of Table \ref{tab:ablation_results} shows that ablating ODs generally leads models to predict a \textit{larger} number of distinct tokens, compared to the corresponding non-ablated runs (column \textit{abl-OD}; stable-12b is an exception). Ablating an equivalent number of non-OD dimensions does not have this effect. Even more strikingly, when keeping only the few ODs in the last layer, the number of distinct tokens that are predicted is drastically reduced (column \textit{only-OD}; cf.\ the contrast with \textit{only-rnd}, that is, when keeping only an equivalent number of random dimensions active).

ODs thus seem to steer models towards predicting a small set of tokens. Table \ref{tab:od_favored_tokens} contains the tokens that are most strongly associated with ODs (they are predicted over 1,000 times when ablating all dimensions except ODs).%
\footnote{To avoid noise, we only consider tokens that were predicted by the full model at least once.}
The table suggests that ODs generally favor very common tokens like \textit{\_the}, \textit{\_and}, or \textit{\_in}, which, as we have already remarked, is a sensible heuristic for a model to encode, since natural language texts are heavily skewed towards a few very frequent types \citep{Baayen:2001}. This might explain why using a few dozen last-layer ODs for prediction generally leads to accuracies that, while low, are non-negligible: for example, \textit{\_the} and \textit{\_a} alone account for 7.5\% of the ground-truth tokens in our dataset.  Again, gemma-9b, stable-12b and opt-13b show different patterns. Gemma-9b has only one token meeting our condition, stable-12b none. Opt-13b has only one OD-favored token which, however, is predicted for almost every input in the only-OD condition.
\begin{table}[htb]
    \centering \small 
    \begin{tabular}{l p{0.33\textwidth}} 
    \toprule
    model & tokens (\# of predictions) \\
    \midrule
    pythia-12b & \_the (15272); \_a (5015); \_D (1716); \\
    mistral-7b & \_the (1665;) \_a (1619); \_ (1568); \_un (1531); \_two (1163); \_large (1106) \\
    llama-8b & \_in (15172); , (10063); \_( (9293); \_and (5227); \_ (1930);\\
    olmo2-13b & \_the (21532); \_ (11275); , (3515); \_The (3181); \_in (1241); \_A (1018); \\
    qwen-14b & \_ (11967;) \_the (5982); , (4708);\\
    opt-13b & I (46181); \\
    gemma-9b & \_ (1762);\\
    stable-12b & n.a. \\
    \bottomrule
    \end{tabular}
    \caption{\small Most strongly OD-favored tokens for each model (predicted over 1K times in only-OD condition, i.e., when ablating all dimensions except ODs). In parentheses: number of times the token is predicted in this condition. No tokens in stable-12b met the criterion. The underscore denotes the space character, which varies by model.}
    \label{tab:od_favored_tokens}
\end{table}

In Appendix~\ref{sec:generation}, we also qualitatively investigate the effect of the ablation on the generative capabilities of the models. As could be expected, we observe that ablating the ODs typically has a significantly larger impact than ablating random dimensions. While in the latter case even removing 3000 dimensions still allows the model to produce meaningful sentences, in the former the ablation of a much smaller number of ODs (even just 5 in the case of qwen-14b) completely disrupts the model's performance. This implies that any generative downstream task will be dramatically affected.

\begin{figure*}[htb]
    \centering
    \subfloat{\includegraphics[scale=0.37]{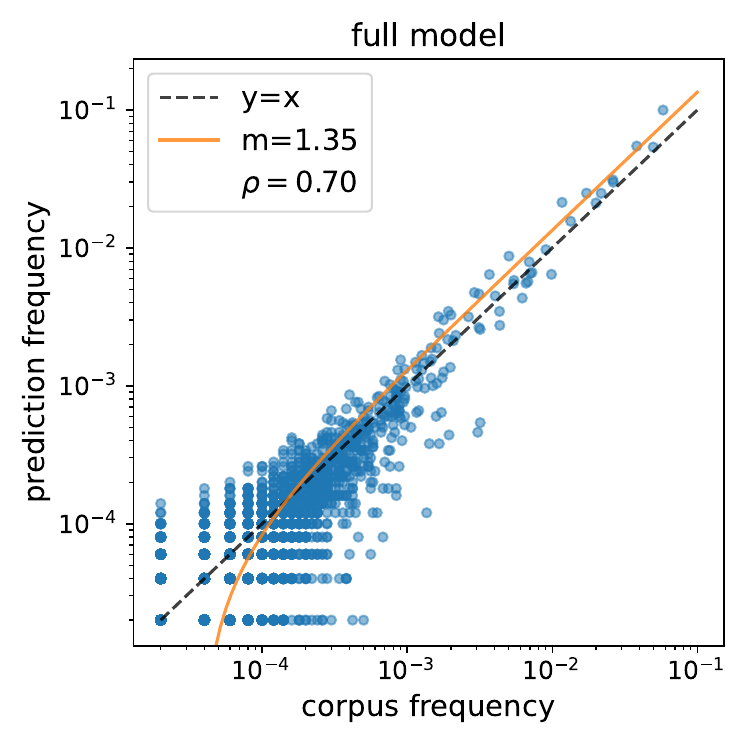}}
    \subfloat{\includegraphics[scale=0.37]{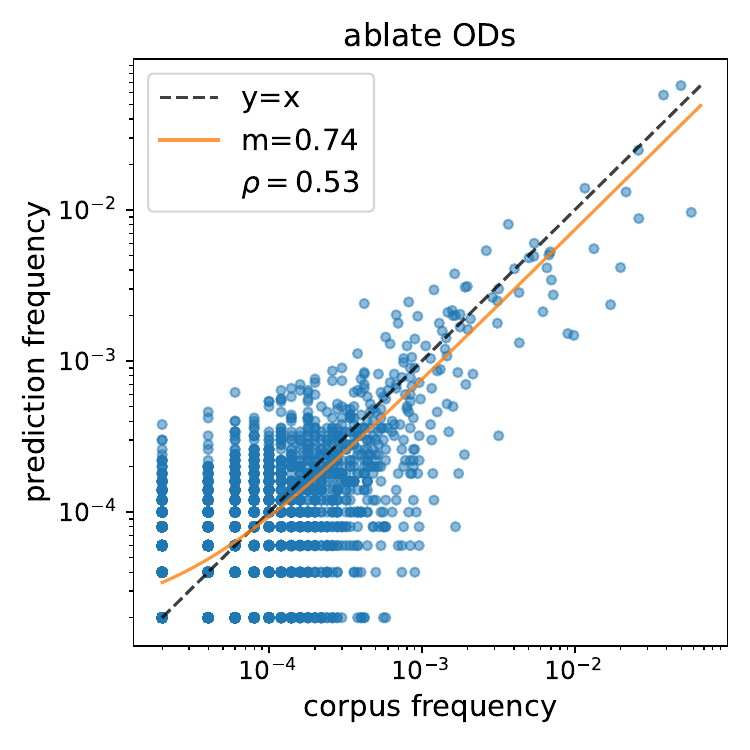}}
    \subfloat{\includegraphics[scale=0.44]{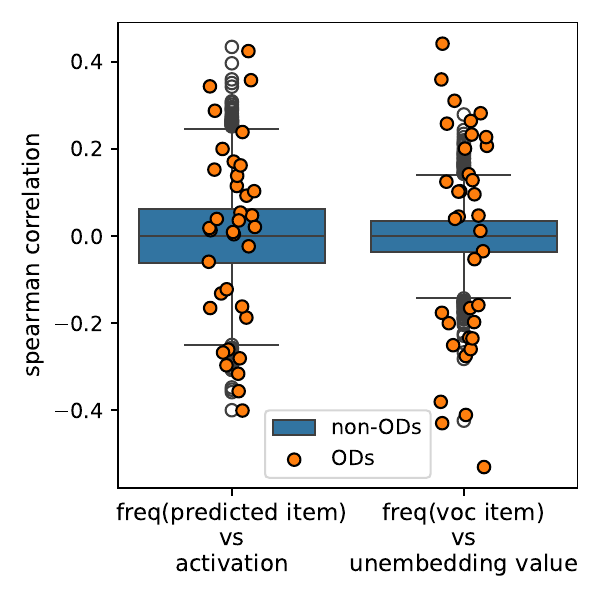}}
    \caption{\small Frequency and ODs in pythia-12b. \textbf{Left and middle panel}: Prediction frequency in function of corpus-estimated frequencies for the full model and the OD-ablated model, in log-log scale. The ablation decreases the frequency for frequent tokens and increases the frequency for rare tokens. \textbf{Rightmost panel, left boxplot}: distribution of the Spearman correlation between the activation value of the last context tokens
    and the corpus-estimated frequency of the model-predicted tokens. \textbf{Right boxplot}: distribution of the Spearman correlation between the values in the unembedding matrix corresponding to a given dimension and the corpus-estimated frequency of corresponding vocabulary items.
    The correlation is computed independently on each dimension. Results for non-ODs are grouped, while ODs are reported as orange dots (scattered along the x-axis for visualization purposes).}
    \label{fig:prediction-corpus-frequency-pythia}
\end{figure*}

\paragraph{Relationship between ODs and frequency} 
The results above, together with the qualitative examples in Table~\ref{tab:od_favored_tokens}, seems to suggest that ODs tend to favor frequent tokens.
To test this hypothesis in a systematic way, we examine the relationship between the frequency with which a model predicts a given token, on the one hand, and its overall frequency (which we estimate with a corpus), on the other.\footnote{Ideally, frequency estimates should be drawn from the LMs' training corpora, but, as we don't have access to those for most models, we use the full WikiText-103 corpus to estimate overall token frequencies.}
We expect OD-ablated models (i.e., models where ODs are set to 0) to decrease the prediction of frequent tokens, and consequently increase the prediction of less frequent tokens, compared to the full models.
The first two panels of Fig.~\ref{fig:prediction-corpus-frequency-pythia} present evidence of this for pythia-12b.
They show the tokens predicted by the full model (left) and the OD-ablated model (middle), sorted by their corpus-estimated frequency and frequency of prediction in each condition, in log-log scale.
They also contain the line of a linear regression fit and its slope, as well as the Spearman correlation coefficient $\rho$.
Both indicators show that ablating ODs indeed affects the relationship between the overall frequency of a token and the times it is predicted. First, whereas the relationship is always positive (models predict more frequent tokens more often than less frequent tokens), the $\rho$ coefficient goes down from 0.7 to 0.53 when ablating ODs, and the slope similarly goes down from 1.35 to 0.74.%
\footnote{The latter also means that the full model over-predicts and the ablated model under-predicts frequent words (and conversely for less frequent words), something that is more clearly shown by the regression lines (perfect correlation with slope 1 would mean that all points are on the $y=x$ line).}

Table \ref{tab:slopes-spearmans-od-frequency} in Appendix \ref{app:ods-and-frequency} confirms that this tendency generalizes: the slope of a linear fit decreases for all models when ODs are ablated, except for stable-12b; as for the correlation coefficient $\rho$, it decreases for all models when ODs are ablated, although the effect is negligible for those models that have already shown some deviant behavior in the previous analyses (opt-13b, stable-12b, and gemma-9b).

\paragraph{Role of ODs in token prediction} 
We next examine the mechanics by which ODs favor frequent tokens in most models. 
First, we hypothesize that the activations of a significant number of individual ODs will be highly correlated with token frequency (either positively or negatively).
The boxplots in the rightmost panel of Fig.~\ref{fig:prediction-corpus-frequency-pythia} provide evidence for pythia-12b. 
The left boxplot shows the distribution of correlations between the corpus frequency of the predicted tokens and the activation profiles of each last-layer dimension across the dataset. This reflects the extent to which, across predicted tokens for our dataset, each dimension tends to be more strongly activated proportionally to the frequency of the predicted tokens. We see that the correlations of the vast majority of non-OD dimensions (summarized in the blue boxplot) cluster around 0, whereas the correlation scores for the ODs (shown as individual orange dots) are much more spread, with a significant proportion exhibiting larger correlations.
Similarly, we find a relation between the corpus frequency of individual vocabulary items and their values in OD vs.~non-OD dimensions of the unembedding matrix, as follows.
The second boxplot in the rightmost panel of Fig.~\ref{fig:prediction-corpus-frequency-pythia} shows the distribution of correlations between the values of each dimension across the unembedding matrix rows and the corpus frequencies of the tokens corresponding to these rows. 
Again, the correlations are generally low for non-ODs, but much more spread for ODs. Given that the output distribution of a language model is obtained by multiplying the unembedding matrix by the last-layer activations, this suggests that ODs will favor more frequent tokens, because the latter tend to have higher unembedding vector values exactly in correspondence to the ODs. The same distributions are plotted for the other models in Fig.~\ref{fig:freqs_vals_all} of Appendix \ref{app:ods-and-frequency}, confirming the same trends across all of them.

These correlations on their own are suggestive, but not enough to uncover the specific mechanics, since, as we just discussed, dimension activations on an LM's last layer contribute to the prediction of a specific token through their dot product with the equivalent dimensions in the unembedding matrix row corresponding to the token. The result of this dot product is the \textit{logit} score for that token: the higher this quantity, the larger the probability that the model will assign to the token. To analyze the interaction between the last layer and the unembedding matrix, we separately measure the contribution of ODs and non-ODs to the logit scores of a given token.
We do that by computing the dot product between the activations of the dimensions of interest and their corresponding unembedding vector values (Appendix \ref{app:logit-contribution-analysis} contains the equations for completeness).

\begin{figure}[tbh]
    \centering
    \subfloat{\includegraphics[scale=0.4]{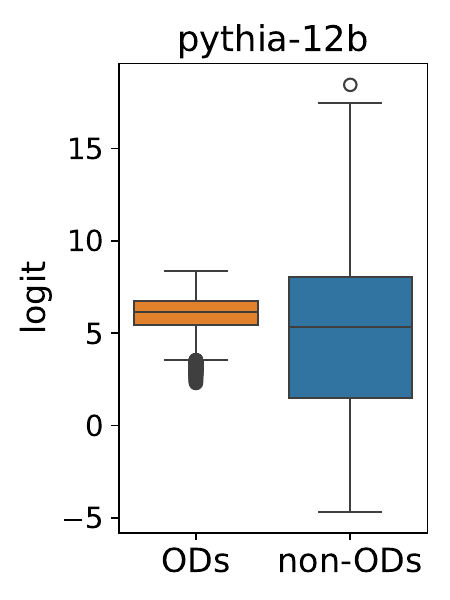}}
    \subfloat{\includegraphics[scale=0.4]{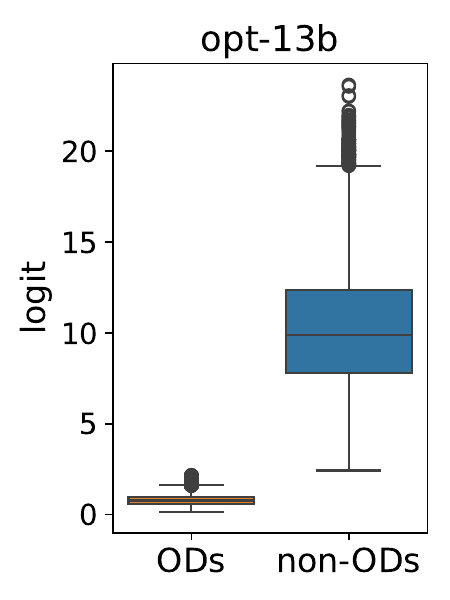}}\\
    \subfloat{\includegraphics[scale=0.4]{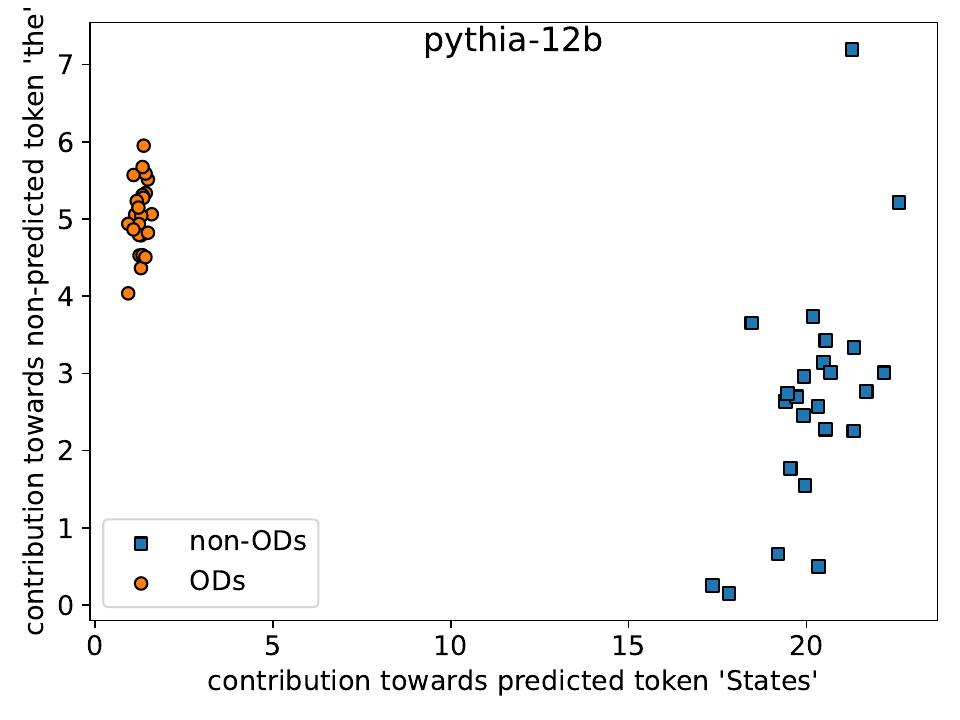}}
    \caption{\small \textbf{Upper:} Distributions of cumulative OD and non-OD logit contributions in all contexts in which an OD-favored token is predicted, for pythia-12b and opt-13b.
    \textbf{Lower}: OD and non-OD logit contributions to OD-neutral \textit{\_States} and OD-favored \textit{\_the} in all contexts in which \textit{\_States} is predicted by pythia-12b. 
    }
    \label{fig:ods_boxplots-logit-contributions-pythia}
\end{figure}

The upper panel in Fig.~\ref{fig:ods_boxplots-logit-contributions-pythia} focuses on OD-favored tokens (listed in Table \ref{tab:od_favored_tokens}), contrasting pythia-12b and opt-13b.
It displays the cumulative logit contributions of ODs and non-ODs to the prediction of each OD-favored token.
We see a different pattern.
In pythia-12b, as expected, the ODs consistently provide a significant positive contribution to predicting the token.
Moreover, this contribution is comparable in size to that of the non-ODs (whose contribution shows a larger variance).%
\footnote{Note that it is not surprising that non-ODs also contribute to promoting OD-favored tokens in contexts where they are appropriate.}
This equivalence is remarkable, as in one case we are looking at the cumulative contribution of 36 ODs, in the other to that of the remaining 5,084 dimensions. Instead, in opt-13b the contribution of the ODs sums to virtually 0. This might go some way towards explaining why this is one of the models for which there is no clear effect of OD ablations, nor strong OD/frequency correlations. While this model does feature a small number of ODs with across-the-board extreme activations, these activations, when multiplied by the unembedding matrix and summed, might cancel each other out, suggesting that, in this case, the model is not really using the ODs to control output prediction.%
\footnote{Recall that, when only ODs are kept, opt-13b almost invariably predicts the single token \textit{I} (Table \ref{tab:od_favored_tokens}). The results in Fig.~\ref{fig:ods_boxplots-logit-contributions-pythia} suggest that this might be more of a ``default'' behavior of the model, than something specifically triggered by the ODs.}
Fig.~\ref{fig:od-favored_all} in Appendix \ref{app:logit-contribution-analysis} shows the OD vs.~non-OD contributions for the other models (except stable-12b, which, as mentioned above, has no OD-favored token according to our criteria).
While gemma-9b behaves like opt-13b, all other models behave like pythia-12b, with a strong contribution of ODs to OD-favored token prediction.

Therefore, in general ODs promote the prediction of frequent tokens, which raises the question: how can a model also predict less frequent tokens?
Remember that ODs, by definition, have extreme values in at least 50\% of the inputs, so they will be strongly active in many cases where it's adequate to predict less frequent tokens.
One possibility is that, while their magnitude is almost constantly large, ODs only provide a \textit{positive} contribution to predicting OD-favored tokens in appropriate contexts, but they give a \textit{negative} contribution to OD-favored token predictions in other contexts, thus penalizing them. A second hypothesis is that ODs always bring about a positive contribution to the prediction of OD-favored tokens, but this is counter-balanced by the total logit mass of the other dimensions when other tokens are more contextually appropriate. To verify which hypothesis is right, we study a set of \textit{OD-neutral} tokens, defined as tokens that a model predicts with equal frequency when the ODs are ablated as when they aren't (e.g.\ \textit{\_times}, \textit{\_command}, \textit{\_States} for pythia-12b; see Table \ref{tab:od_neutral_tokens} in Appendix \ref{app:logit-contribution-analysis} for the full set).
We sample maximally 10 such tokens per model, and we limit the choice to tokens that are predicted at least 10 times by the full model. 
Now, for each context in which the model predicted (that is, assigned the largest probability to) one of these tokens, we contrast the OD and non-OD contributions towards the logit of the relevant token, on the one hand, and the logits of the already mentioned OD-favored tokens, on the other, to find by which means the OD-neutral token ``won'' against the OD-favored rivals.

The lower panel of Fig.~\ref{fig:ods_boxplots-logit-contributions-pythia} illustrates the analysis with the case of the OD-neutral token \textit{\_States}.
The plot contains two data points for each context in which pythia-12b predicts \textit{\_States}, one corresponding to the logit contributions of ODs (orange dots) and one to those of non-ODs (blue squares). 
On the x-axis, the logit values for the token \textit{\_States} are displayed, while the y-axis shows the logit values for the OD-favored token \textit{\_the}.
If the negative-contribution hypothesis was right, the ODs would provide a negative contribution to the \textit{\_the} prediction in this case, as it is a context in which it predicts \textit{\_States}. However, this is not what we find; we see that ODs always strongly favor \textit{\_the} (and only provide a modest positive contribution to \textit{\_States}). Indeed, the OD logit contribution towards \textit{\_the} here is comparable to the average OD contribution towards OD-favored tokens when the latter are chosen (around 5, cf. the upper left boxplot of the figure). The reason that \textit{\_the} is ultimately not chosen in the \textit{\_States} contexts is that the sum of all non-ODs gives a large positive contribution to the \textit{\_States} prediction, that surpasses the cumulative weight of the ODs in favor of \textit{\_the}. 

In Fig.~\ref{fig:od-neutral_all} in Appendix~\ref{app:logit-contribution-analysis}, we show that this pattern generalizes across OD-neutral and OD-favored tokens.
In all models, the OD-neutral tokens are selected because of the logits of the non-ODs; OD logits either present higher values for OD-favored tokens or are neutral, with values near 0 (the latter is the case for gemma-9b and opt-13b).

We conclude that the models do not learn to modulate ODs so that they are positive in contexts where OD-favored tokens should be predicted, and negative otherwise. Rather, ODs act like a constant term providing a positive logit contribution towards OD-favored tokens, and it is the job of the other dimensions to learn to counterbalance this contribution when other tokens should be predicted. We can think of ODs as implementing a baseline heuristic to always predict frequent tokens, with the rest of the dimensions working their way around them to provide more appropriate context-dependent predictions.

\paragraph{Which model parameters boost OD activations?}
The observation that ODs consistently exhibit large values across most inputs, combined with evidence that these values are not strongly context-modulated, suggests that OD activations result directly from large model weights boosting them. Given that most ODs appear in the last layer only, we consider three plausible OD boosters: the down-projection matrix of the last MLP (mapping MLP-internal representations to last-layer activations), the weight terms of the final LayerNorm, and the bias terms of the final LayerNorm, for models that have these parameters (see Table \ref{tab:MBHyper} in Appendix \ref{app:language-model-details}).\footnote{In Appendix \ref{app:od-boosting}, we comment on the negative results we obtained concerning the last layer's components of the attention heads.} 
Adopting a methodology similar to \cite{Cancedda:2024}, we assess the possible contribution of the MLP matrix to the ODs by first factorizing it with a singular value decomposition. We then examine whether the dominant (i.e., first few) singular vectors' ``spikes'' (dimensions with values exceeding three standard deviations from the mean) tend to align with the last layer's ODs. 
See Appendix \ref{app:od-boosting} for a detailed explanation of the procedure. Likewise, for the weights and biases of the LayerNorm, we similarly define outlying spikes and, again, we check whether ODs tend to coincide with them. 
A coincidence between spikes and ODs would suggest that the OD values are influenced by the principal directions in the last-layer MLP down-projection matrix, as well as by elevated biases and weights in the final layer normalization.

Results for pythia-12b are presented in Fig.~\ref{fig:pythia-12b_eigenLayerNorm}.%
\footnote{Here and in Fig.~\ref{fig:OtherModels_eigenLayerNorm} of Appendix \ref{app:od-boosting}, we only report the first 4 MLP singular vectors, but in all cases similar patterns are observed for a much broader set of dominating singular vectors; see Fig.~\ref{fig:linerar_combination} in Appendix~\ref{app:od-boosting}.}
The overlap of spikes in the the MLP matrix and ODs is remarkable; for instance, 4 out of 7 spikes in the first singular vector are in ODs, as are 25 out of 43 for the second singular vector. 
These overlaps have \( p \approx 0 \) of occurring by chance, based on simulating a random overlap distribution. Similarly, the right panels of Fig.~\ref{fig:pythia-12b_eigenLayerNorm} show that ODs are also boosted by the LayerNorm weight and bias parameters: 16 out of 52 spikes in the weight are in ODs, as are 20 out of 39 for the bias parameter. 
To sum up, in pythia-12b ODs are boosted by both the MLP matrix and the LayerNorm weights and biases.
Fig.~\ref{fig:OtherModels_eigenLayerNorm} in Appendix \ref{app:od-boosting}
confirms a similar pattern across most models.
Yet, for llama-8b and opt-12b the boost appears to originate solely from the MLP, while for stable-12b the effect is particularly evident in the weights.

\begin{figure*}[htb]
    \centering
    \includegraphics[width = 400pt]{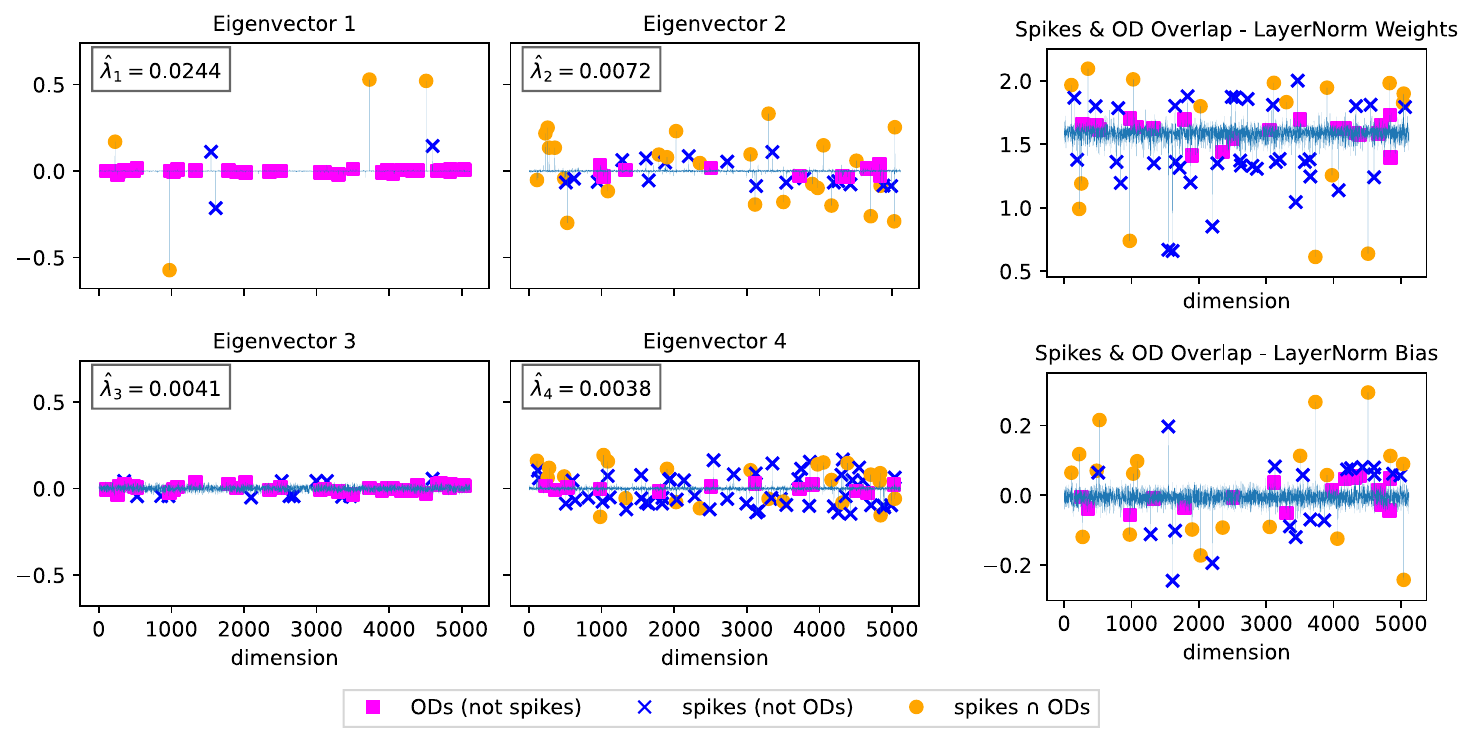}
    \caption{\small \textbf{Left:} Top-4 singular vectors and values of the last-layer MLP down-projection matrix of pythia-12b. \textbf{Right:} Final LayerNorm weight and bias for pythia-12b. Spikes in parameter values that correspond to ODs are visualized as orange circles.
    }
    \label{fig:pythia-12b_eigenLayerNorm}
\end{figure*}

Overall, we find that there are fixed parameters that enhance ODs, and this boosting can be encoded in both the last MLP down-projection matrix and the LayerNorm parameters, highlighting the importance of looking for multiple parametric sources of extreme values.

\paragraph{ODs' emergence and role during training}
As a last experiment, we looked at the emergence of the ODs and the effect of their ablation during the stages of training of pythia-12b.
A significant number of ODs appears around steps 3000/4000. From here onward, ablating ODs impacts accuracy and leads to the prediction of more varied tokens, suggesting that ODs started assuming the function of frequent token boosters. Interestingly, early in training (step 500), there are virtually no ODs, yet even the full model predicts relatively few tokens. The full results are reported in Table~\ref{tab:training_full} in Appendix~\ref{app:training}.
If we examine which tokens are over-predicted at this stage compared to their actual dataset frequencies,\footnote{To reduce noise, we only consider tokens occurring at least 100 times as ground-truth predictions.} we find them to include \textit{\_the, \_that, \_and, \_not, \_a, \_be, \_of}. This indicates that the model has learned right from the start that predicting very frequent tokens pays off.
In later steps, the model specializes the ODs to implement this heuristic, allowing the remaining units to focus on context-aware prediction, resulting in more varied outputs and improved accuracy.

\section{Discussion and conclusion}
\label{sec:discussion}

We  studied a class of LM extreme values we dubbed \textit{outlier dimensions} (ODs). Unlike the more widely studied ``massive activations'' \citep{sun2024massiveactivationslargelanguage,An_Zhao_Yu_Tang_Wang_2025}, 
ODs exhibit extreme activations consistently across a range of inputs, rather than in response to specific ones. 
Unlike massive activations, 
which influence model output indirectly through attention sinking, 
ODs directly influence model predictions by favoring frequent tokens independently of context. As such, they function as a hard-coded module developed 
to address an important characteristic of text: its extremely skewed distributions. 
Interestingly, \citet{Stolfo:etal:2024} found a set of frequency-modulating neurons, defined as entries in an MLP internal representation, in several smaller models. Future work should connect their results to ours, to check if we are examining different aspects of the same mechanism.

Five out of eight LMs we studied have ODs linked to frequent-token predictions. Divergent results for the other models suggest, however, that the development of frequent-token predicting ODs, while a common strategy discovered by many models, is by no means necessary. All divergent models have units that meet our OD criteria, but they differ in other ways: in particular, opt-13b and gemma-9b have fewer ODs than the other models, and OD ablation does not impact their accuracy, whereas stable-12b has a larger number of ODs, and ablating them does affect performance. When only ODs are retained, opt-13b almost always predicts the same token (\textit{I}), whereas gemma-9b only mildly favors the space token, and stable-12b has no clearly favored token under this ablation. 
Overall, considering that for opt-13b and gemma-9b the logit contribution of ODs approaches 0 even when predicting their OD-favored tokens, it appears that ODs may serve a different function in these LMs, potentially akin to that of attention-sinking massive activations. However, we defer a more thorough characterization of ODs in these divergent models, as well as of the properties that made them diverge, to future studies.

\section*{Acknowledgments}
We thank Beatrix Miranda Ginn Nielsen, Santiago Acevedo, Diego Doimo, Javier Ferrando, Alessandro Laio and the members of the UPF COLT group for feedback. Our work was funded by the European Research Council (ERC) under the European Union’s Horizon 2020 research and innovation programme (grant agreement No. 101019291). We also received funding from the Catalan government (AGAUR grant SGR 2021 00470). NG also received the support of a fellowship from Fundación Ramón Areces. 
GB also received the support of grant PID2020-112602GB-I00/MICIN/AEI/10.13039/501100011033, funded by the Ministerio de Ciencia e Innovación and the Agencia Estatal de Investigación (Spain).
This paper reflects the authors’ view only, and the funding agencies are not responsible for any use that may be made of the information it contains.
\section*{Limitations}
\begin{itemize}
    \item As stated at the beginning of our manuscript,  our analysis focuses exclusively on outliers in the last hidden layer, since their direct influence on the next token prediction can be clearly observed. Although outliers also exist in earlier layers, assessing their impact and linguistic role is more complex and less straightforward.
    \item Our analysis of the ODs at training time is limited to a single model, pythia-12b, for which early training checkpoints are available. Consequently, our findings may not generalize to other models. We were unable to test this further because the only other publicly available model with training checkpoints is olmo2-13b; however, its earliest checkpoint is not early enough in training to capture initial OD behavior, as fully-trained-model OD behavior is already firmly in place by then.
\end{itemize}

\bibliography{colm2025_conference,marco}
\newpage
\appendix

\section{ODs across layers}
\label{app:ods-layers}

\begin{figure*}[htb]
    \centering
    \subfloat{\includegraphics[scale=0.4]{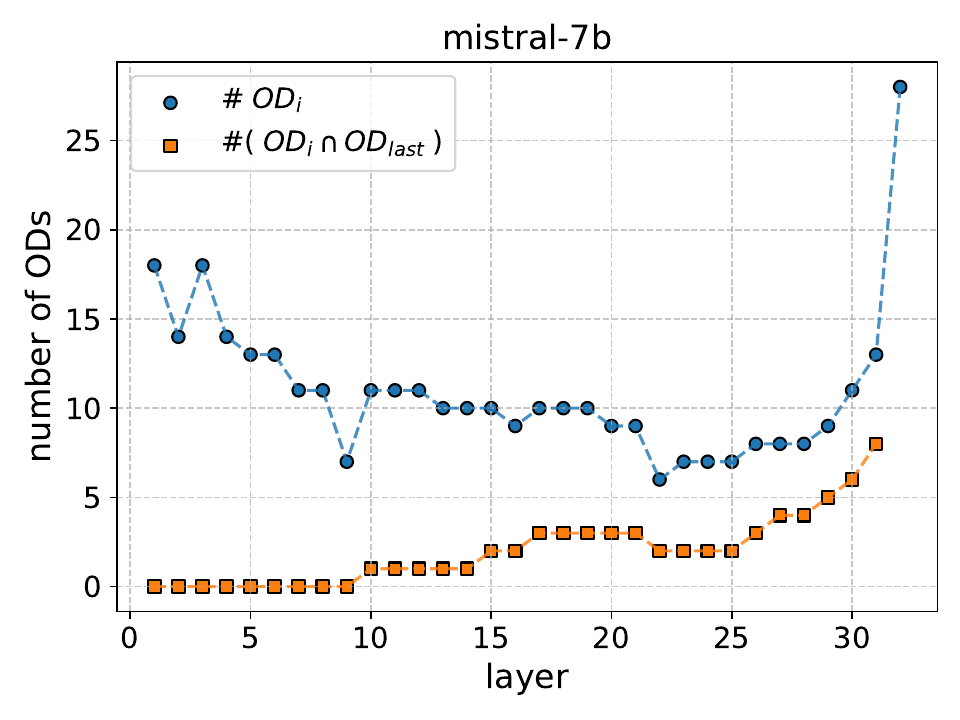}}%
    \subfloat{\includegraphics[scale=0.4]{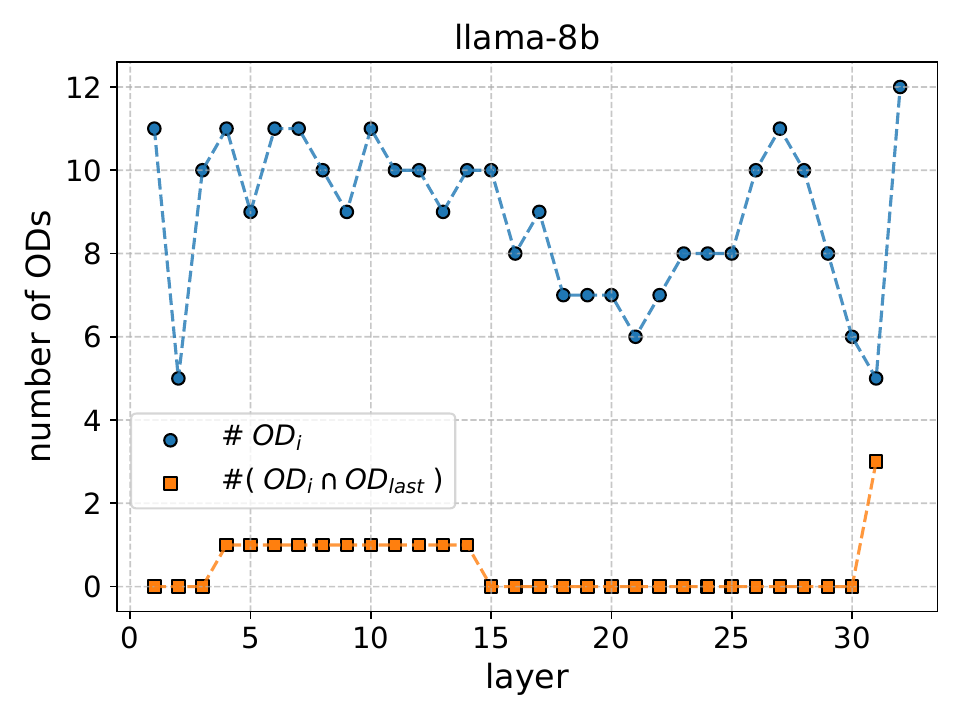}}\\
    \subfloat{\includegraphics[scale=0.4]{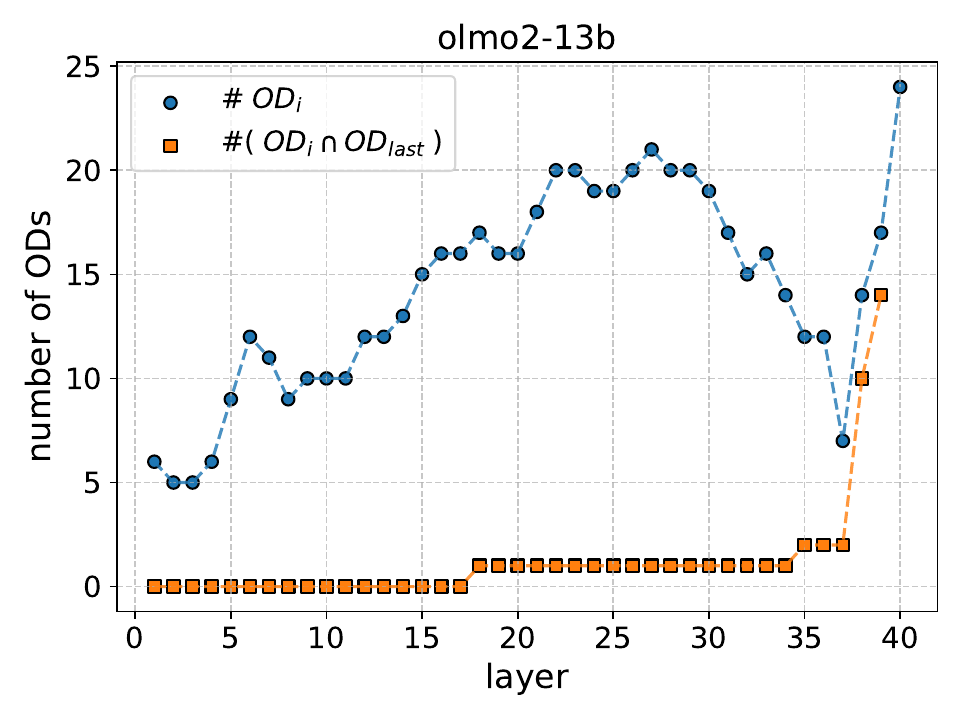}}%
    \subfloat{\includegraphics[scale=0.4]{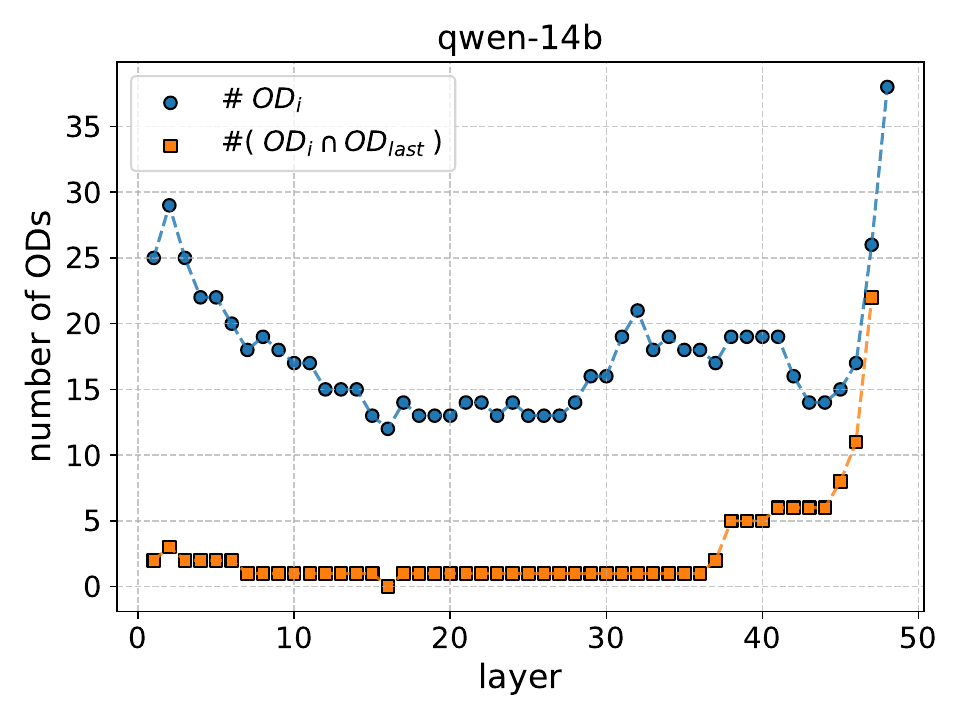}}\\
    \subfloat{\includegraphics[scale=0.4]{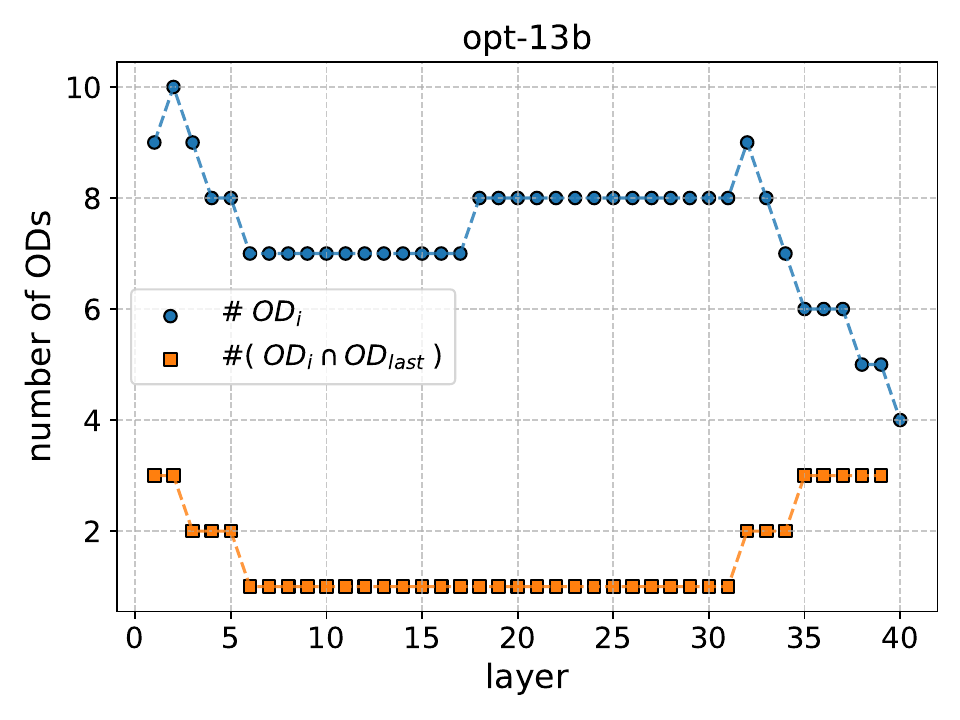}}%
    \subfloat{\includegraphics[scale=0.4]{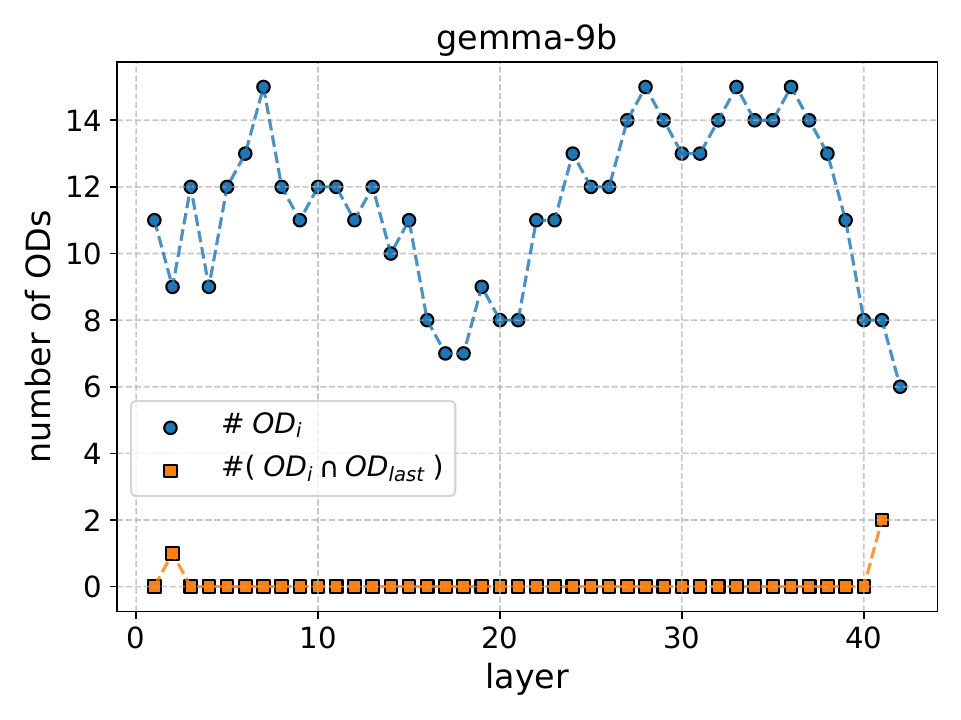}}\\
    \subfloat{\includegraphics[scale=0.4]{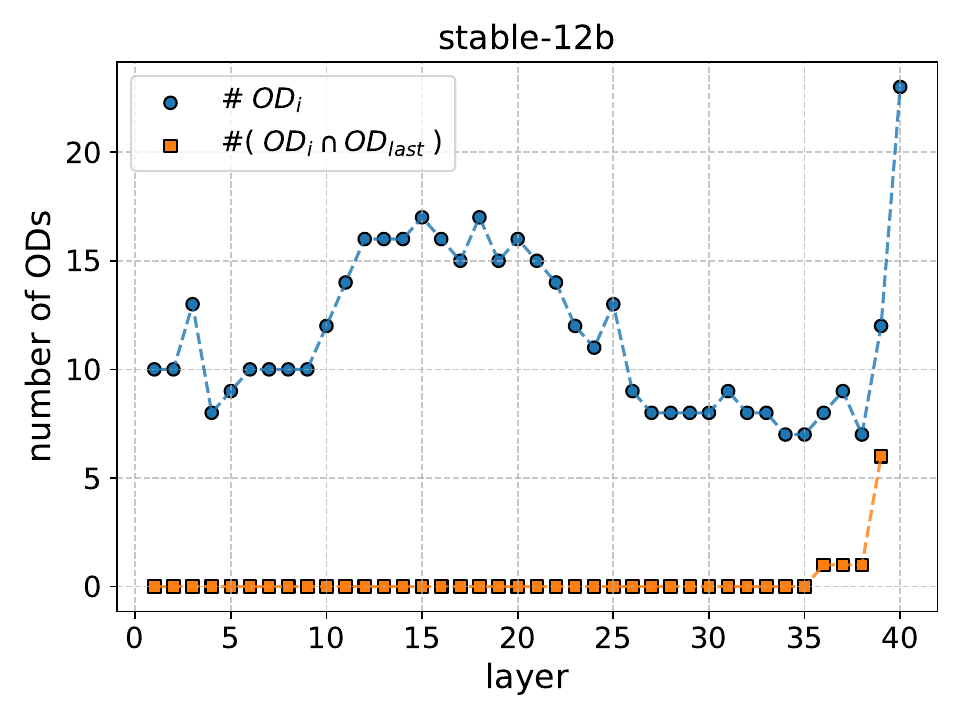}}
    \caption{\small Number of ODs across layers (blue dots) and number of ODs in each layer that are also present in the last layer (orange squares). The last point for the overlap is not reported as it coincides with the actual number of ODs.}
    \label{fig:ods_overlap_all}
\end{figure*}

Fig.~\ref{fig:ods_overlap_all} reports the number of per-layer ODs (as identified by our criterion) and the number of ODs that also appear in the last layer for all studied models except pythia-12b, whose equivalent data are shown on the right panel of Fig.~\ref{fig:ods} in the main text. We observe that in all models, but gemma-9b and opt-13b, the last layer has many more ODs than the previous ones. Moreover, we also see how the ODs present in the last layer tend to appear only towards the end, with different onset curves for the various models. Togethes, these observations suggest that the last layer has a special behavior that differs from that of the previous ones. 

\section{Language model details}
\label{app:language-model-details}

Table \ref{tab:MBHyper} summarizes the main characteristics of the language models we used. 

\NewDocumentCommand{\rot}{O{35} O{1em} m}{\makebox[#2][l]{\rotatebox{#1}{#3}}}%

\begin{table*}
\centering
\begin{adjustbox}{
width=\textwidth}
\centering
\small 
\ttfamily
\begin{tabular}{lllllllll} 
\toprule
model name & \rot{\textbf{Llama}-3-8B}  & \rot{\textbf{Mistral}} \rot{-7B-v0.3}  & \rot{\textbf{gemma}-2-9b} & \rot{\textbf{stablelm}} \rot{-2-12b} & \rot{\textbf{pythia}-12b} \rot{-deduped}& \rot{\textbf{opt}-13b}  & \rot{\textbf{OLMo}-2} \rot{-1124-13B} & \rot{\textbf{Qwen}2.5} \rot{-14B}\\ 
\hline
\hfill\\
size: & 8B & 7B & 9B & 12B &  12B & 13B & 13B & 14B\\ 
number of layers: & 32 & 32  & 42 &  40  & 36  & 40  &  40 & 48\\ 
hidden dimension: & 4096 & 4096 & 3584  & 5120  &5120&5120&5120&5120\\ 
\# attention heads: & 32 & 32 & 16 &  32 & 40 &  40 & 40 &  40\\ 
\# key/value heads: & 8 &8 & 8 & 8&n.a  & n.a & n.a & 8\\ 
head size: & n.a & 128 & 256 & n.a & n.a & n.a &  n.a & 128 \\ 
FFN size: & 14336 & 14336 & 14336 & 13824 & 20480 &20480  & 13824 & 13824 \\ 
vocab size: & 128256 & 32768 & 256128 & 100352 &  50688 & 50272  & 100352  & 152064 \\ 
position embeddings: & \checkmark &\checkmark  &  \checkmark& \checkmark & \checkmark & \checkmark & \checkmark & \checkmark \\ 
max pos. embeddings: & 8192 & 8192 & 8192 & 4096 & 2048 & 2048  &  4096 &  131072 \\ 
tied word embeddings: & $\times$  &$\times$   & \checkmark & $\times$  & $\times$  & \checkmark  & $\times$  &  $\times$ \\ 
distilled model: & $\times$  & $\times$  & \checkmark & $\times$  & $\times$  & $\times$  & $\times$  &  $\times$ \\ 
activation: & SwiGLU & SwiGLU & GeGLU & SwiGLU  & GeLU  & ReLU  & SwiGLU  &  SwiGLU\\ 
LayerNorm (LN) type: & RMSNorm & RMSNorm & RMSNorm&  LN & LN& LN & RMSNorm  & RMSNorm \\ 
LN has bias parameter:  & $\times$  & $\times$  & $\times$   & \checkmark &\checkmark   & \checkmark &  $\times$  & $\times$  \\
torch dtype: & bfloat16 &  bfloat16& float32  & bfloat16 & float16  & float16& float32 & bfloat16 \\
\hfill \\
\hline
\hfill \\
public training data:  & $\times$  & $\times$  &  $\times$ & \checkmark  &\checkmark   & $\times$  &   \checkmark &  $\times$  \\
checkpoints available: & $\times$  & $\times$ &  $\times$  & $\times$    & \checkmark & $\leq$ 10 & \checkmark   &  $\times$ \\
\hfill \\
\bottomrule
\end{tabular}
\end{adjustbox}
\caption{\small Configuration details of the pre-trained LMs we experimented with; n.a. means \textit{not available} information.}\label{tab:MBHyper} 
\end{table*}

\section{Surprisal effects of ablations}
\label{app:surprisal}

Table \ref{tab:surprisal} reports the 1st, 2nd (median) and 3rd quartiles for the distribution of surprisal in the different ablation modalities. For each context, the surprisal is calculated as $S=-\log(p)$, where $p$ is the probability attributed by the model to the ground truth next token.
These observations broadly confirm the results obtained for accuracy (see Table \ref{tab:ablation_results} in the main text). Note that for stable-12b, one of the divergent models, only-OD surprisal is larger than only-random surprisal: we leave to future studies a better understanding of how ODs affect output probabilities for this LM.

\begin{table*}[htb]
    \centering
    \begin{tabular}{lcccc}
        \toprule
        \multicolumn{5}{c}{Surprisal (1st, 2nd, 3rd quartiles)} \\
        \midrule
        Model & FM & ablate-OD & only-OD & only-random \\
        \midrule
        pythia-12b & 0.6, 1.9, 4.0 & 1.0, 2.8, 4.9 & 8.0, 9.0, 9.8 & 10.7 \\
        mistral-7b & 0.5, 1.6, 3.5 & 0.6, 1.9, 4.0 & 8.2, 8.9, 9.6 & 10.3 \\
        llama-8b & 0.4, 1.5, 3.4 & 0.6, 2.1, 4.2 & 8.5, 9.7, 10.7 & 11.7 \\
        olmo2-13b & 0.2, 1.2, 3.2 & 0.6, 2.2, 4.6 & 6.4, 7.8, 9.4 & 11.4 \\
        qwen-14b & 0.3, 1.4, 3.4 & 0.7, 3.2, 7.0 & 8.8, 10.6, 15.1 & 11.8 \\
        opt-13b & 0.6, 1.9, 4.1 & 0.6, 2.0, 4.1 & 10.3, 10.5, 10.7 & 10.8 \\
        gemma-9b & 0.0, 1.4, 14.9 & 0.0, 1.5, 14.9 & 11.3, 12.1, 13.3 & 12.4 \\
        stable-12b & 0.4, 1.5, 3.4 & 0.8, 3.3, 8.4 & 22.4, 27.8, 34.2 & 11.4 \\
        \bottomrule
    \end{tabular}
    \caption{1st, 2nd and 3rd quartiles of the distribution of surprisal, computed on the predicted token for each input, for each ablation type. The large values of the 3rd quartile hint at highly positive-skewed distributions. For the only-random case we report only the median as the other quantiles are identical within the used significant digits. The ablate-random condition is not reported since, as it occurred for accuracy, the values are identical to the ones of full model up to the reported precision. In the only-random case, one can appreciate how the surprisal corresponds to a random guess: by exponentiating these values, one  approximately recovers the vocabulary size.}
    \label{tab:surprisal}
\end{table*}

\section{ODs and frequency}
\label{app:ods-and-frequency}

Table \ref{tab:slopes-spearmans-od-frequency} reports the linear-fit slopes and Spearman correlation coefficients of the relation between generic corpus-estimated frequency of tokens and their prediction frequency in our dataset, for the full models and their OD-ablated versions. For both measures, we observe a drop from full-model to ablate-OD, except for the usual divergent models: opt-13b and gemma-12b essentially show no ablation effect, and for stable-12b we actually observe an \textit{increase} in slope from full-model to OD-ablated.

\begin{table*}[htb]
    \centering
    \setlength{\tabcolsep}{4.5pt} 
    \begin{tabular}{l|cc|cc}
    \toprule
         & \multicolumn{2}{c}{Slope} & \multicolumn{2}{c}{Spearman} \\
         \midrule
         Model & FM & ablate-OD & FM & ablate-OD \\
    \midrule
    pythia-12b & 1.35 & 0.74 & 0.70 & 0.53 \\
    mistral-7b & 1.23 & 1.06 & 0.71 & 0.64 \\
    llama-8b   & 1.26 & 0.86 & 0.71 & 0.58 \\
    olmo2-13b  & 1.24 & 0.64 & 0.74 & 0.59 \\
    qwen-14b   & 1.22 & 0.95 & 0.71 & 0.52 \\
    opt-13b    & 1.34 & 1.25 & 0.70 & 0.67 \\
    gemma-9b   & 1.30 & 1.26 & 0.72 & 0.71 \\
    stable-12b & 1.27 & 1.72 & 0.71 & 0.70 \\
    \bottomrule
    \end{tabular}
    \caption{Linear fit slope and Spearman correlation coefficient for the relation between corpus-estimated and prediction frequencies for the full models and under OD ablation.}
    \label{tab:slopes-spearmans-od-frequency}
\end{table*}

\begin{figure*}[tb]
    \centering
    \subfloat{\includegraphics[scale=0.45]{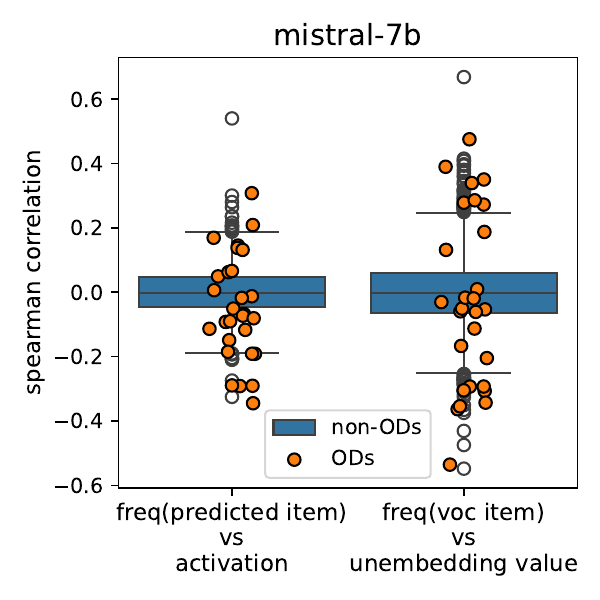}}%
    \subfloat{\includegraphics[scale=0.45]{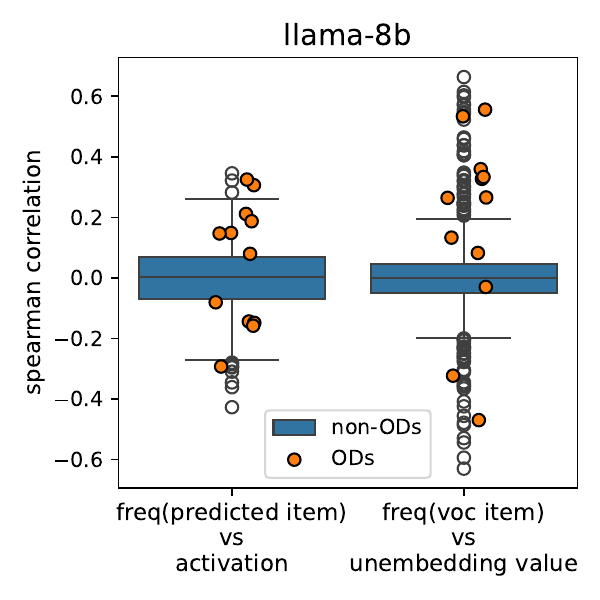}}%
    \subfloat{\includegraphics[scale=0.45]{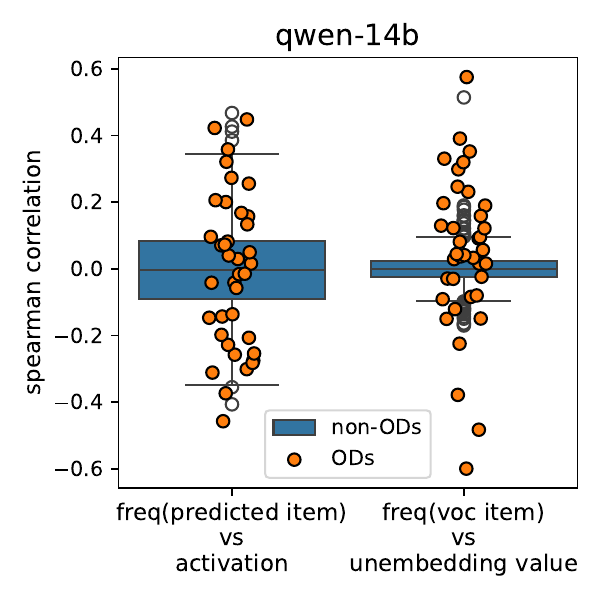}}\\
    \subfloat{\includegraphics[scale=0.45]{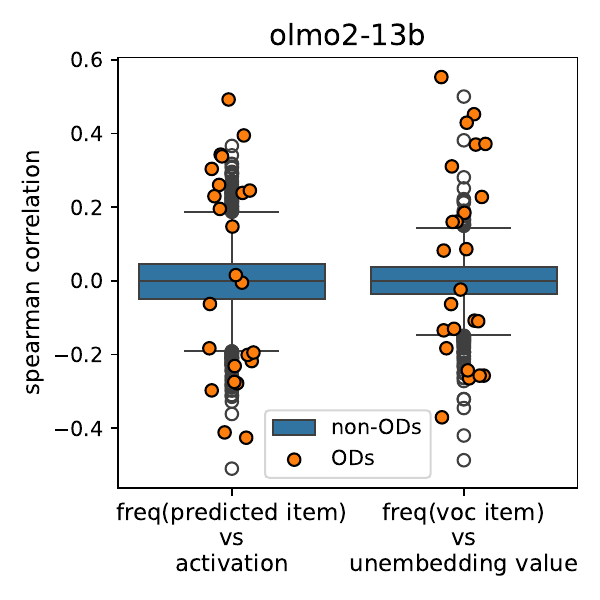}}%
    \subfloat{\includegraphics[scale=0.45]{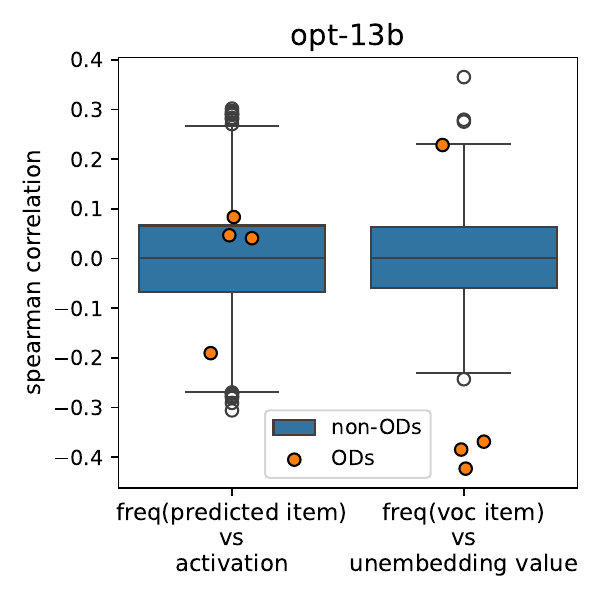}}%
    \subfloat{\includegraphics[scale=0.45]{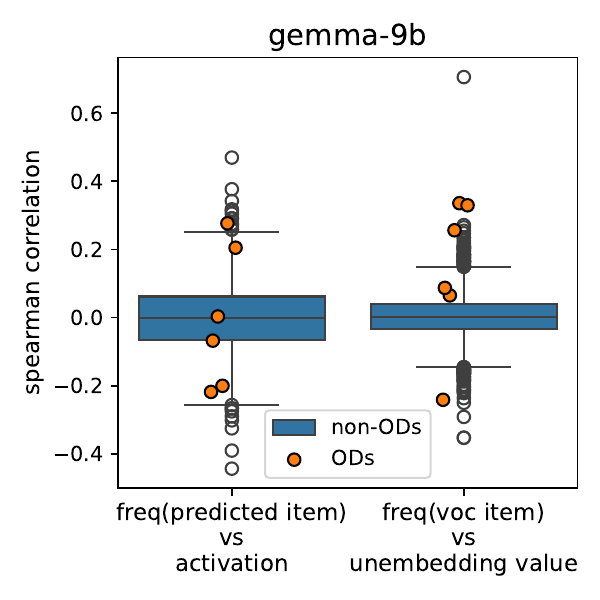}}\\
    \subfloat{\includegraphics[scale=0.45]{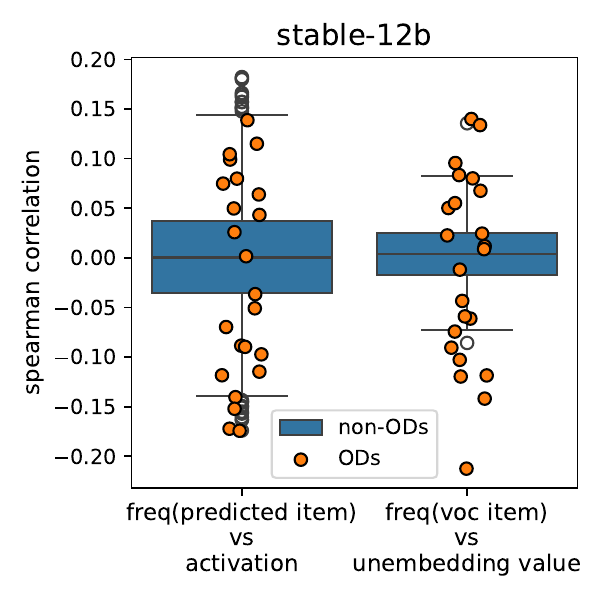}}
    \caption{For each dimension, we compute the Spearman correlation coefficient between: (left) the last-layer activation value of each input context and the frequency of the predicted token; (right) the value in the unembedding matrix of each vocabulary item against its corpus frequency. We group in a boxplot the results for the non-ODS, while explicitly report each OD correlation score as orange circles. Smearing along the x-axis is to help visualization.}
    \label{fig:freqs_vals_all}
\end{figure*}

The left boxplots in Fig.~\ref{fig:freqs_vals_all} present, for all LMs except pythia-12b, the distribution of correlations between the corpus-estimated frequencies of predicted tokens given the inputs in our dataset and the activations of ODs and non-ODs in the last layer of the last context token. The right plots show the distribution of correlations between the corpus-estimated frequencies of the output vocabulary tokens and the OD and non-OD values in the unembedding matrix vectors corresponding to those vocabulary tokens. The same data are shown for pythia-12b in the rightmost panel of Fig.~\ref{fig:prediction-corpus-frequency-pythia} of the main text. Also for the other models we observe that ODs display a much larger probability of showing a higher (absolute) value of the correlation than non-ODs in both scenarios. So, being an OD often implies a large correlation, but the opposite is not always true, as there are many dimensions for which such correlation is high that are not ODs. It could be interesting to see how the models behave upon ablating (or keeping) only those non-ODs with a high correlation coefficient, but we leave this analysis to future work.

\section{Logit contribution analysis}
\label{app:logit-contribution-analysis}

\subsection{Logit contribution computation}

Given a context, an autoregressive neural language model produces a probability distribution over possible next tokens by multiplying the last-layer activations of the last token of the context by the unembedding matrix, and converting the resulting \textit{logit} scores to probabilities through the softmax function. We ignore the latter here, since it is monotonic and we are only interested in comparing which of several possible next tokens gets the largest score. We thus focus on the logit values of the candidate next tokens we are interested in. The logit for candidate next token $t$, denoted $logit_t$, is given by the dot product between the context activation vector $\mathbf{c}$ and the unembedding matrix row corresponding to $t$, $\mathbf{u_t}$:

$$logit_t = \mathbf{c} \cdot \mathbf{u_t} =  \sum_{i \in D} c_i u_{ti}$$

\noindent{}where $D$ is the set of indices of dimensions of the context and unembedding row vectors.

In order to assess how much the ODs contribute to this score, we can simply partition the sum above into two components, one given by the sum of terms corresponding to the dimensions that are in the OD set, and one by the sum of terms for the leftover dimensions. Formally, if $O$ is the set of indices of OD dimensions, the contribution of the ODs to $logit_t$ is:

$$logit_t^{\in OD} = \sum_{i\in{O}} c_i u_{ti}$$

The contribution of the non-OD dimensions is trivially given by:

$$logit_t^{\not\in OD}= logit_t - logit_t^{\in OD}$$

\subsection{Logit contributions in OD-favored predictions}

\begin{figure*}[htb]
    \centering
    \subfloat{\includegraphics[scale=0.35]{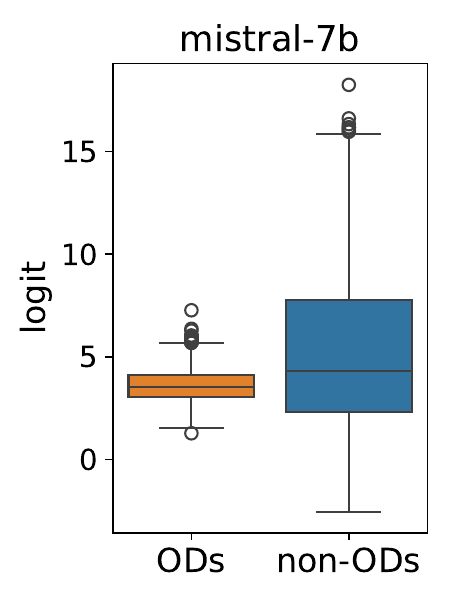}}%
    \subfloat{\includegraphics[scale=0.35]{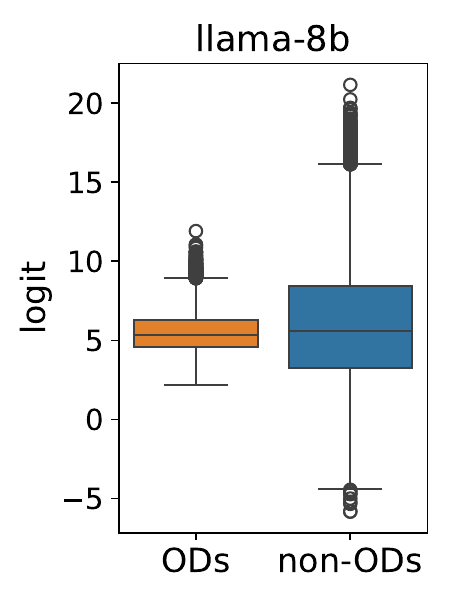}}%
    \subfloat{\includegraphics[scale=0.35]{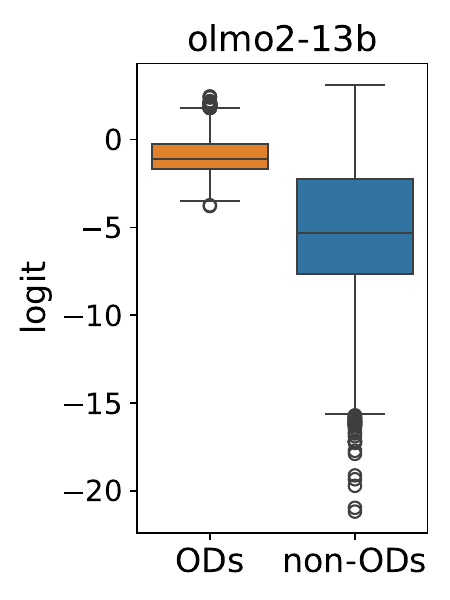}}%
    \subfloat{\includegraphics[scale=0.35]{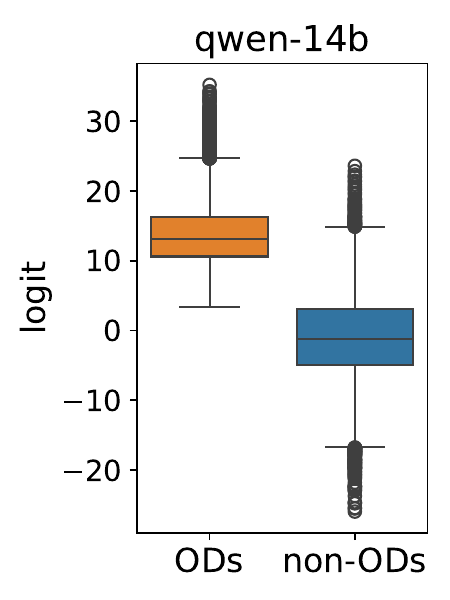}}%
    \subfloat{\includegraphics[scale=0.35]{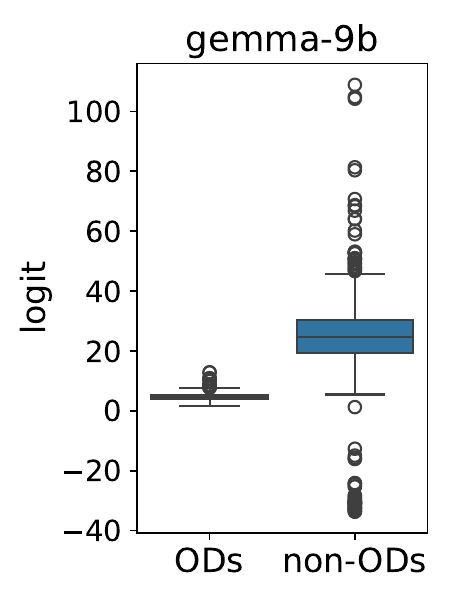}}
    \caption{OD and non-OD logit contributions for all contexts predicting OD-favored tokens}
    \label{fig:od-favored_all}
\end{figure*}

Fig.~\ref{fig:od-favored_all} shows the distribution of logit OD and non-OD contributions towards the prediction of OD-favored tokens for all models except pythia-12b and opt-13b, that are shown in the upper panel of Fig.~\ref{fig:ods_boxplots-logit-contributions-pythia} of the main text. Note that stable-12b is missing because we did not identify any OD-favored token for this model. For all models but opt-13b and gemma-9b, ODs, despite being very few, contribute at least as strongly as non-ODs to token prediction. It is worth remarking that for olmo2-13b, while the ODs are giving the larger contribution to the logits, both contributions are negative.

\subsection{Logit contributions in OD-neutral predictions}

We consider a set of OD-neutral tokens. These are predicted at least 10 times in our dataset by the non-ablated model, and they are predicted the same number of times when ODs are ablated. For qwen-14b, as there was only one token meeting this criterion, we sampled tokens whose OD-ablated/full-model prediction count ratio was between 0.9 and 1.1. We sample maximally 10 tokens per model, shown in Table \ref{tab:od_neutral_tokens}.

\begin{table*}[tb]
    \centering  
    \begin{tabular}{l l} 
        \toprule
        Model & Tokens \\ 
        \midrule
        pythia-12b & \_period \_before \_times \_command \_bridge \_mm \_States \_II \\
        mistral-7b & \_km \_match \_life \_line \_book \_species \_States \_Tour \_season \_City\\
        llama-8b & \_November \_players \_Court \_ships \_species \_Council \_Division \_office \_largest \_Road \\
        olmo2-13b & \_North \_York \_Union \_ships \_money \_Japanese \_success \_times \_ship \_sold\\
        qwen-14b & \_Street \_match \_gave \_York \_interview \_each \_into \_based \_began \_Japan \\
        opt-13b & \_York \_University \_interview \_feet \_water \_part \_ships \_Cup \_men \_half \\
        gemma-9b & \_Road \_feet \_long \_won \_relationship \_mm \_support \_United \_little \_local \\
        stable-12b & n.a. \\ 
        \bottomrule
    \end{tabular}
    \caption{ OD-neutral tokens for the different models. We consider a token OD-neutral if it is predicted from the same representations before and after ablating ODs. A subsample of 10 was extracted if more were available.}
    \label{tab:od_neutral_tokens}
\end{table*}

\begin{figure*}[htb]
    \centering
    \subfloat{\includegraphics[scale=0.42]{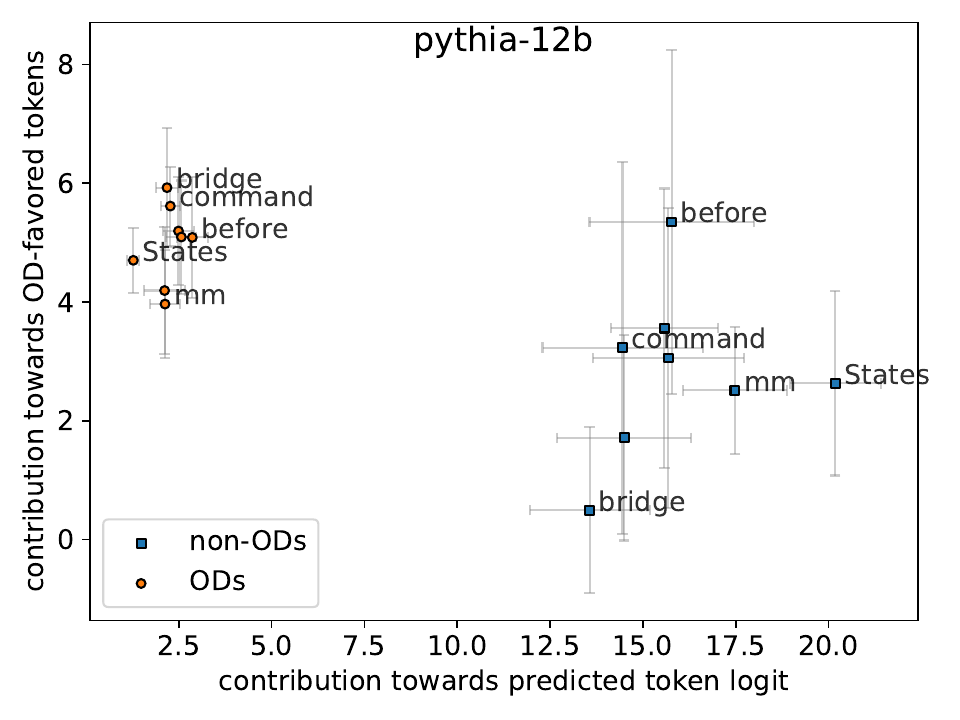}}
    \subfloat{\includegraphics[scale=0.42]{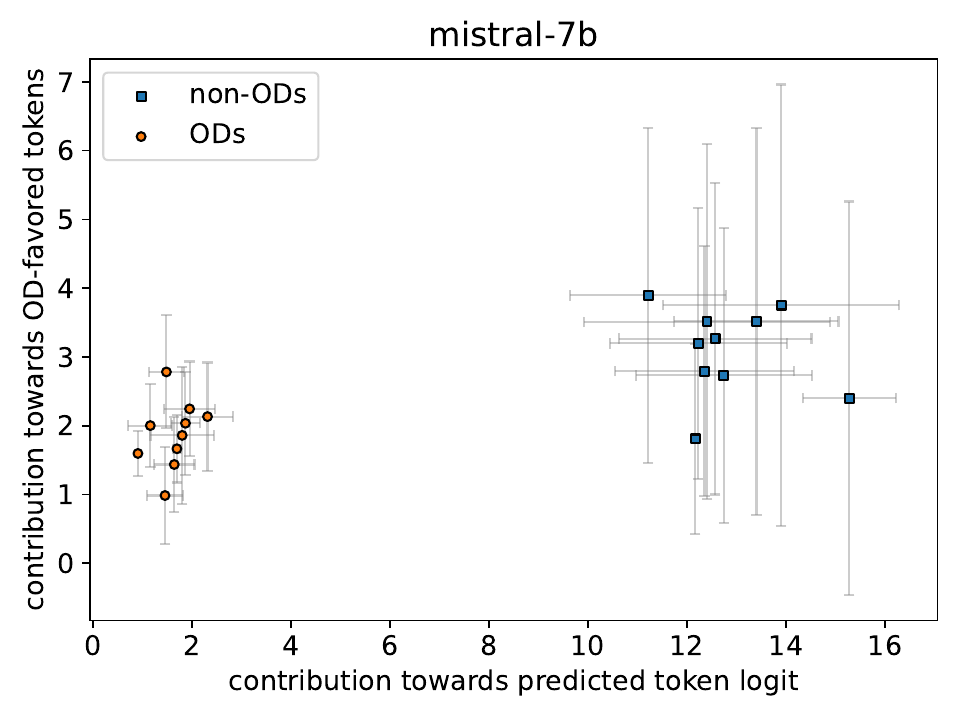}}\\
    \subfloat{\includegraphics[scale=0.42]{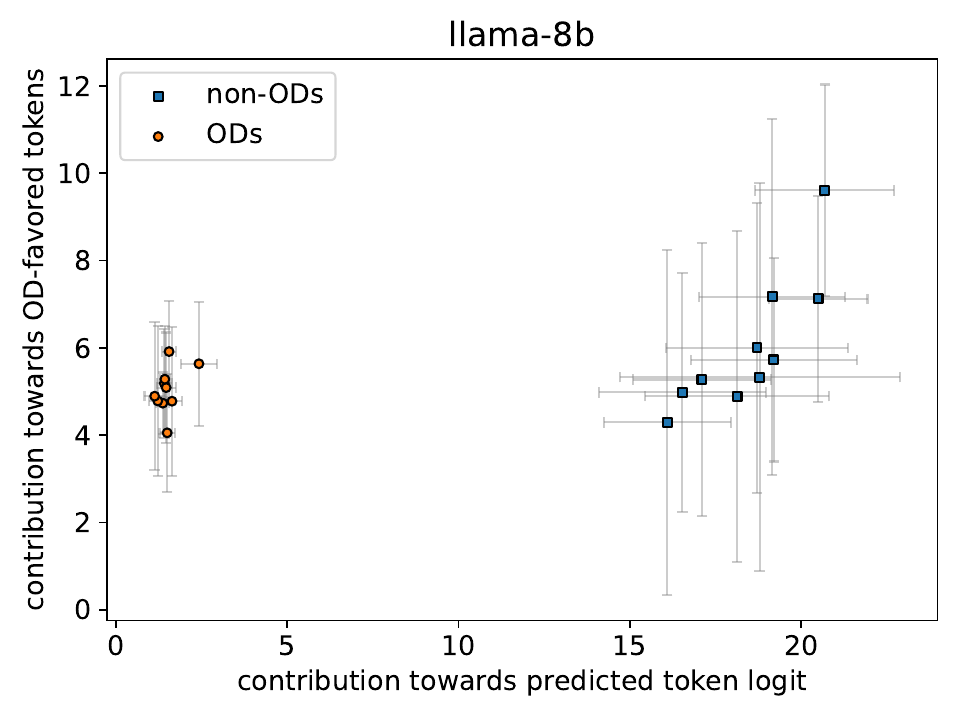}}
    \subfloat{\includegraphics[scale=0.42]{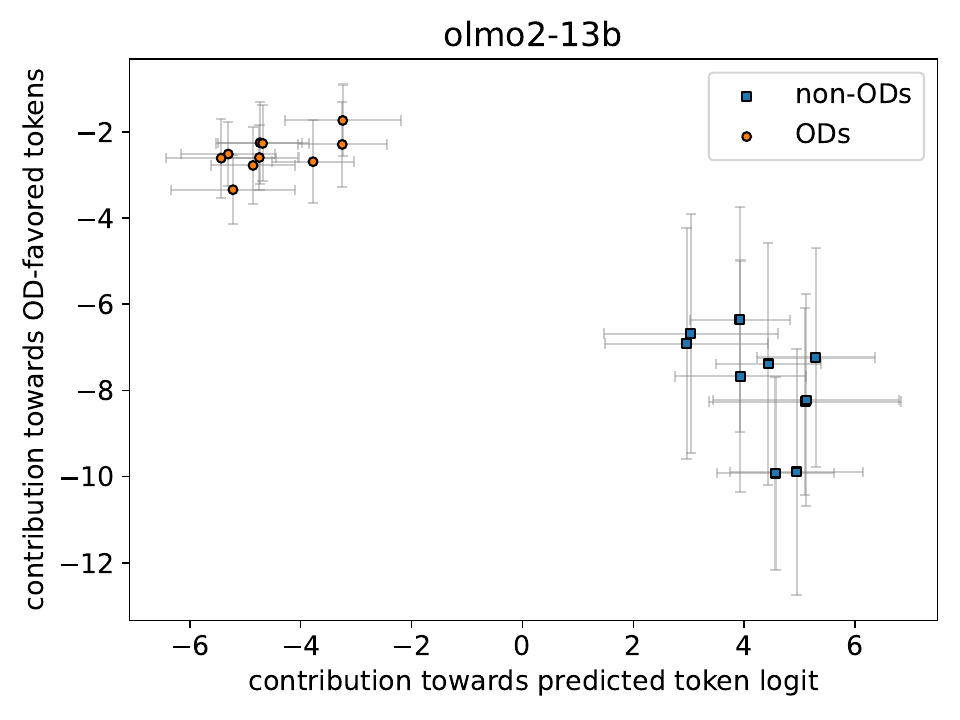}}\\
    \subfloat{\includegraphics[scale=0.42]{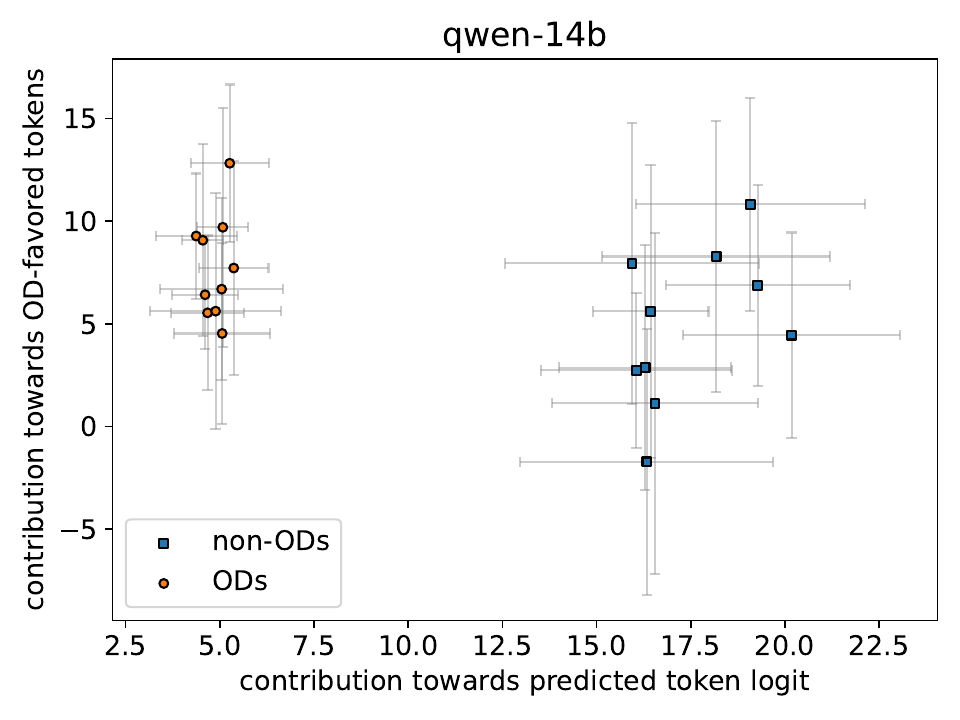}}
    \subfloat{\includegraphics[scale=0.42]{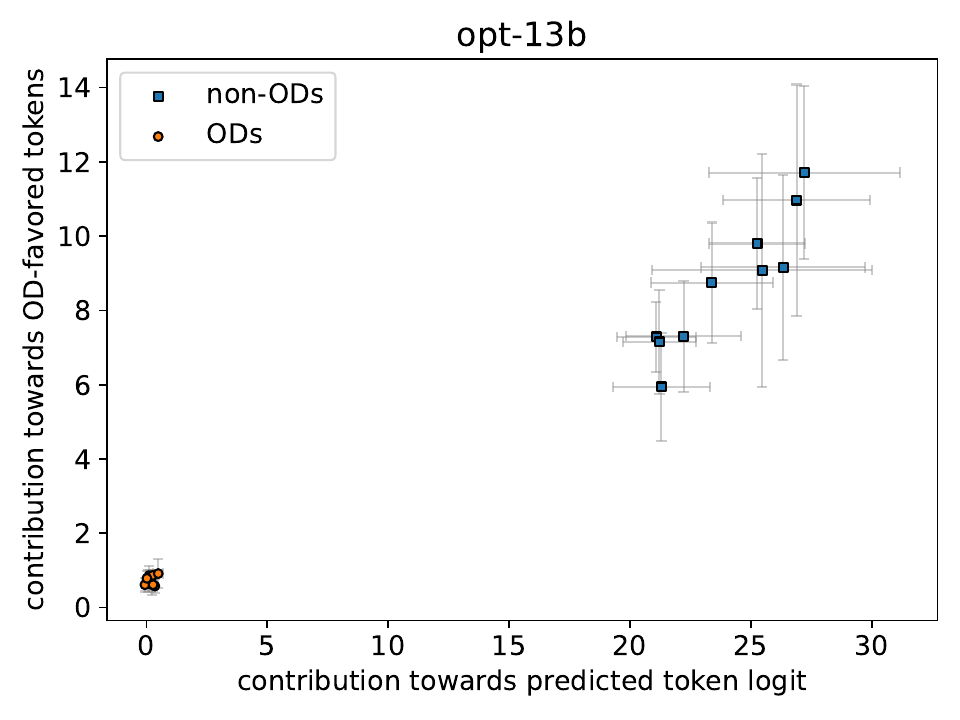}}\\
    \subfloat{\includegraphics[scale=0.42]{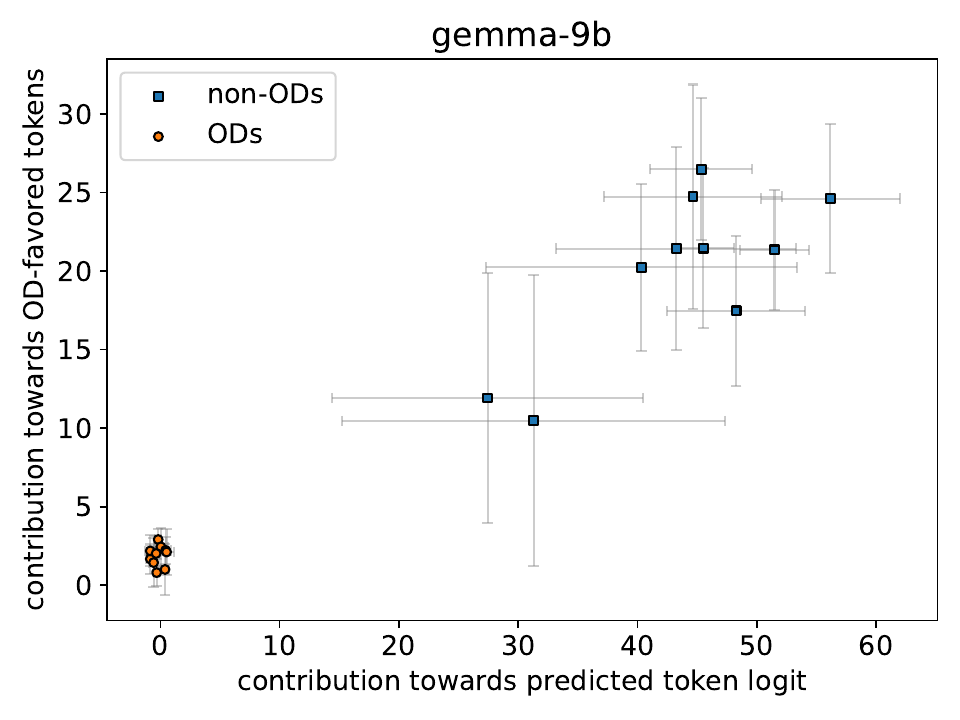}}
    \caption{For each model: OD and non-OD contributions towards OD-neutral and OD-favored tokens in contexts where OD-neutrals are predicted, averaged across the contexts of each OD-neutral token and across contributions to each OD-favored token. We report standard deviations as error bars for clarity.}
    \label{fig:od-neutral_all}
\end{figure*}

Fig.~\ref{fig:od-neutral_all} plots the contribution of ODs and non-ODs towards OD-favored and OD-neutral tokens, in contexts in which the model assigned the larger probability to the latter (for pythia-12b, the same data in the specific case of \textit{\_States} against \textit{\_the} are showed in the lower panel of Fig.~\ref{fig:ods_boxplots-logit-contributions-pythia} in the main text; stable-12b, again, is missing because it has no OD-favored tokens according to our criterion). The contributions are averaged across the contexts of each OD-neutral tokens. For pythia-12b, mistral-7b, llama-8b, olmo-13b and qwen-14b, ODs contribute more strongly to the OD-favored logits, but this effect is outdone, in all cases, by the non-ODs. Note that, as already remarked with respect to the previous figure, for olmo2-13b OD values are negative for both OD-favored and OD-neutral tokens, but closer to 0 for OD-favored, confirming the general trend. For opt-13b and gemma-9b, the OD contributions always sum to values close to 0, confirming the divergent behavior of these models.

\section{OD-boosting parameter analysis}
\label{app:od-boosting}

We explore if ODs are aligned with directions that are boosted by the last MLP down-projection matrix $X$, of shape $d\times h$---where $d$ is the hidden (or embedding) dimension of the model and $h$ is the high-dimensional space between the MLP up- and down-projection matrices (see \texttt{FFN size} in Tab.~\ref{tab:MBHyper}). We then compute the SVD decomposition of $ X = U \Sigma V^T$, where $V$ is a $h\times h$ matrix, $\Sigma$ is a $d\times h$ diagonal rectangular matrix whose entries are called singular values, and $U$ is a $d\times d$ matrix, that ultimately projects an input vector back into the hidden space. We thus focus on the columns of $U$, which are called left singular vectors and have the same dimensionality $d$ of the states from which we extract the ODs.

\begin{figure*}[tb]
\centering
\subfloat{\includegraphics[scale=0.55]{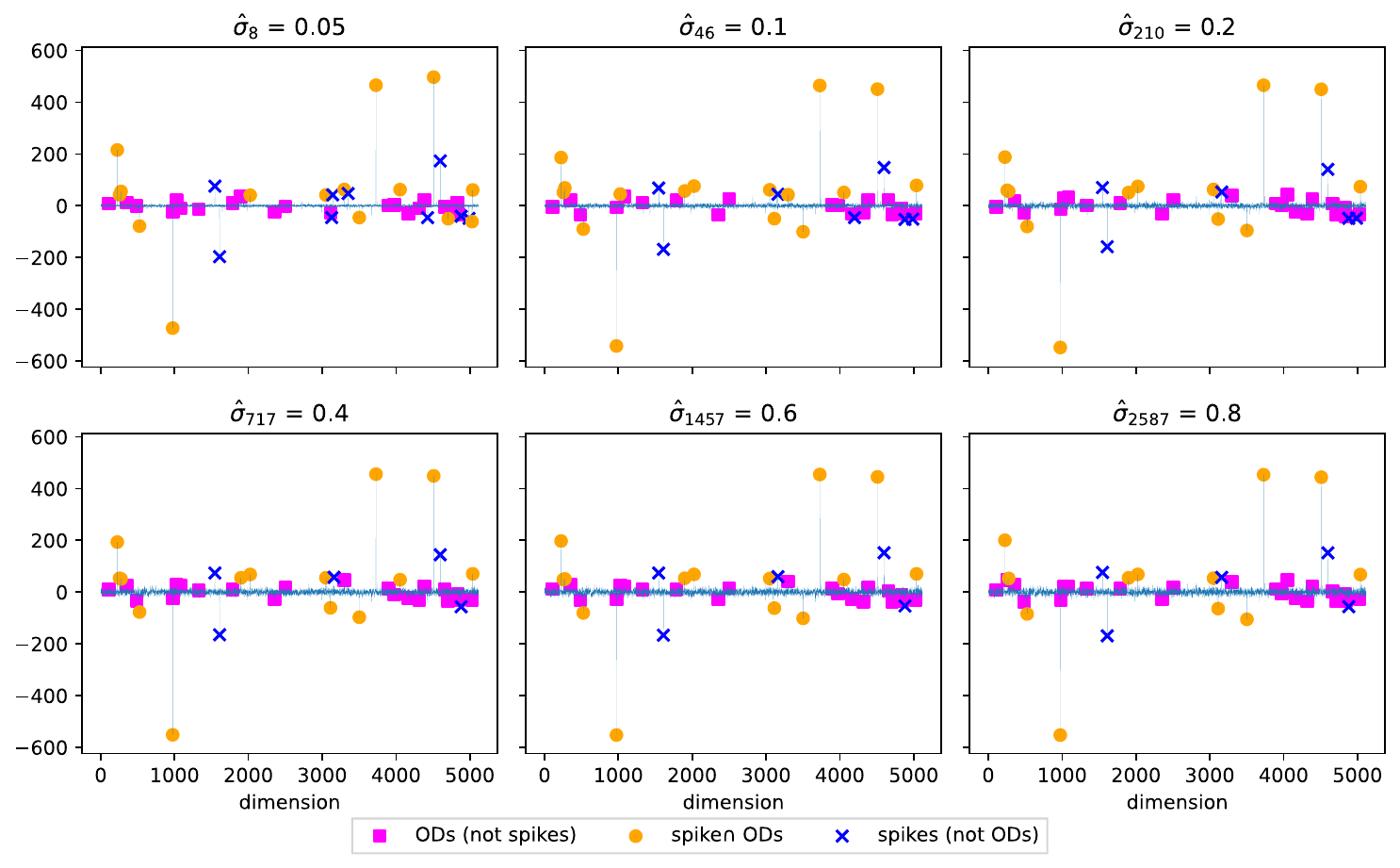}}
\caption{Overlaps between ODs and spikes of the vector obtained as the linear combination of the top N singular vectors, such that cumulative fraction of explained variance is $\hat{\sigma}_N=\frac{\sum_{i=1}^N\lambda_i}{\sum_{i=1}^D\lambda_i}$, where $D$ is the total number of dimensions. Spikes which correspond to ODs are visualized as orange circles.}
\label{fig:linerar_combination}
\end{figure*}

Fig.~\ref{fig:linerar_combination} visualizes vectors given by the linear combination of the top-N singular vectors that account for a certain proportion of variance in the original matrix, weighted by the corresponding singular values, together with the distribution of ODs and ``spike'' dimensions. Here and below, we define as spikes those values of a vector that  lie more than 3 standard deviations away from the mean. The figure confirms the tendency for at least some ODs to coincide with the spikes.

\begin{figure*}%
\subfloat[mistral-7b]{\includegraphics[scale=0.3]{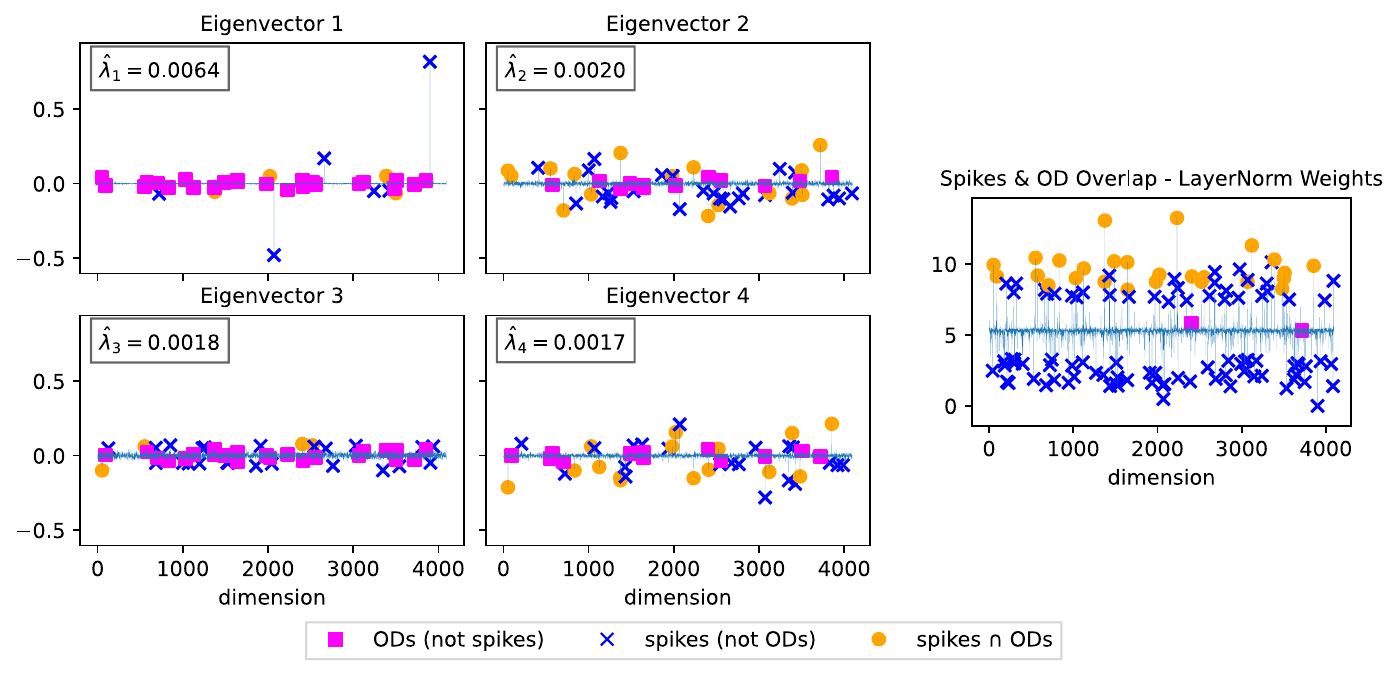}}
\subfloat[llama-8b]{\includegraphics[scale=0.3]{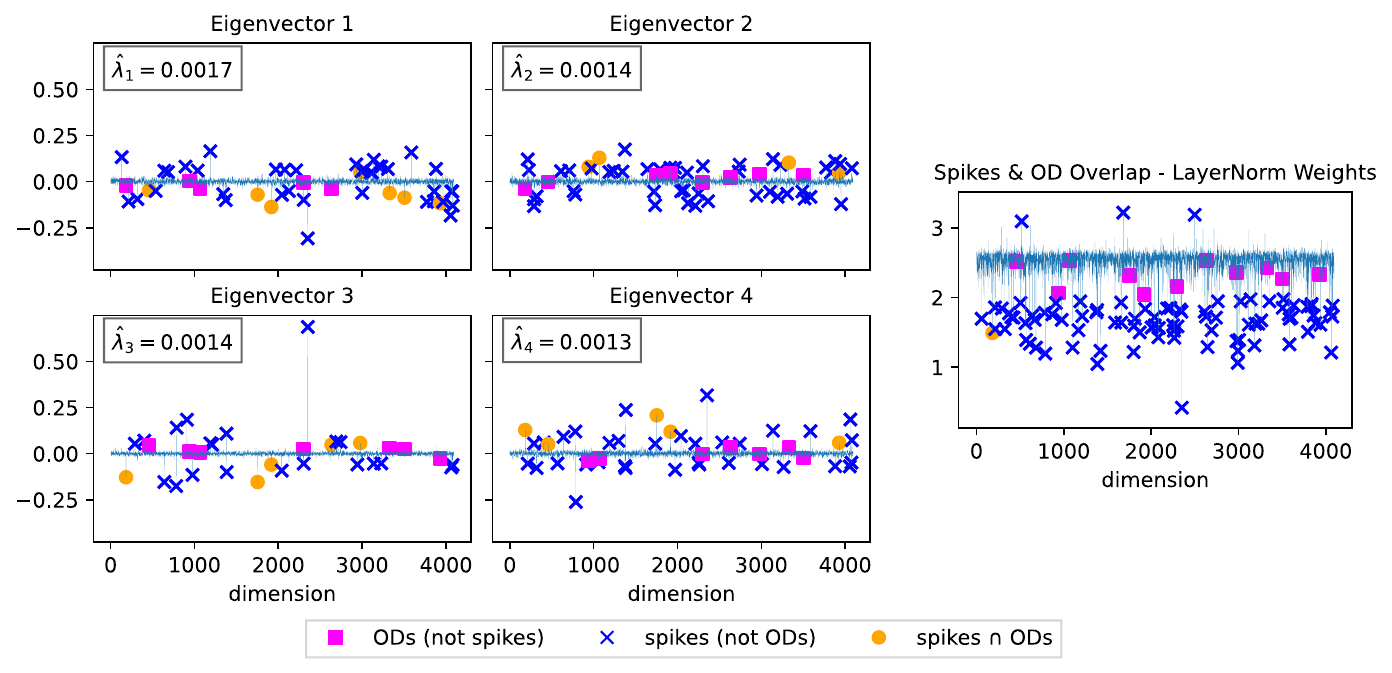}}

\subfloat[olmo-13b]{\includegraphics[scale=0.3]{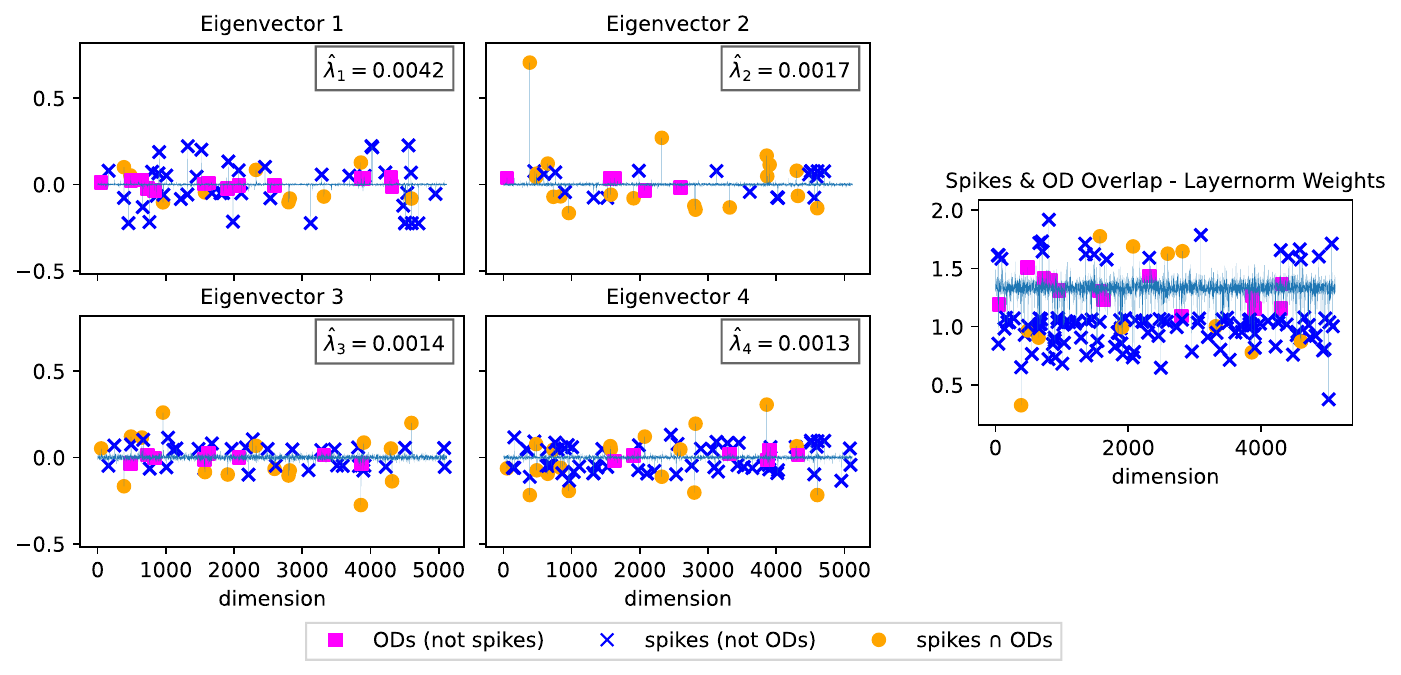}}
\subfloat[qwen-14b]{\includegraphics[scale=0.3]{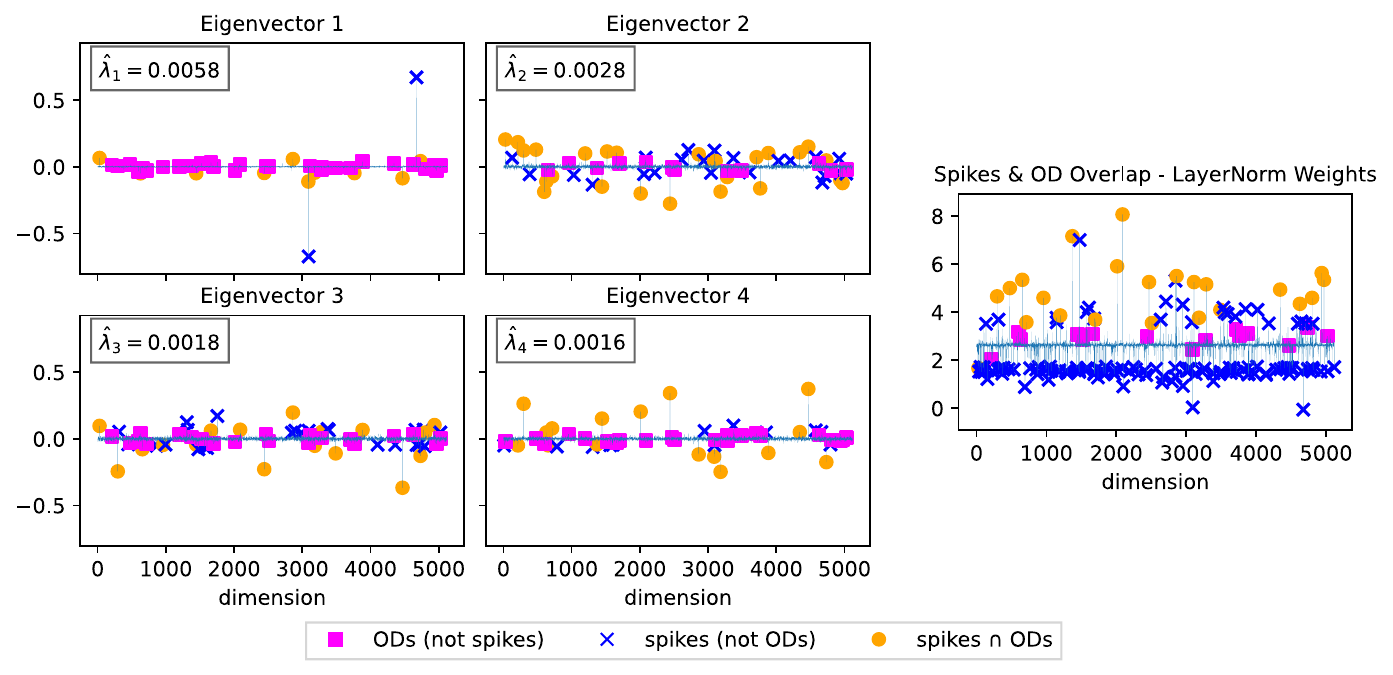}}

\subfloat[opt-12b]{\includegraphics[scale=0.285]{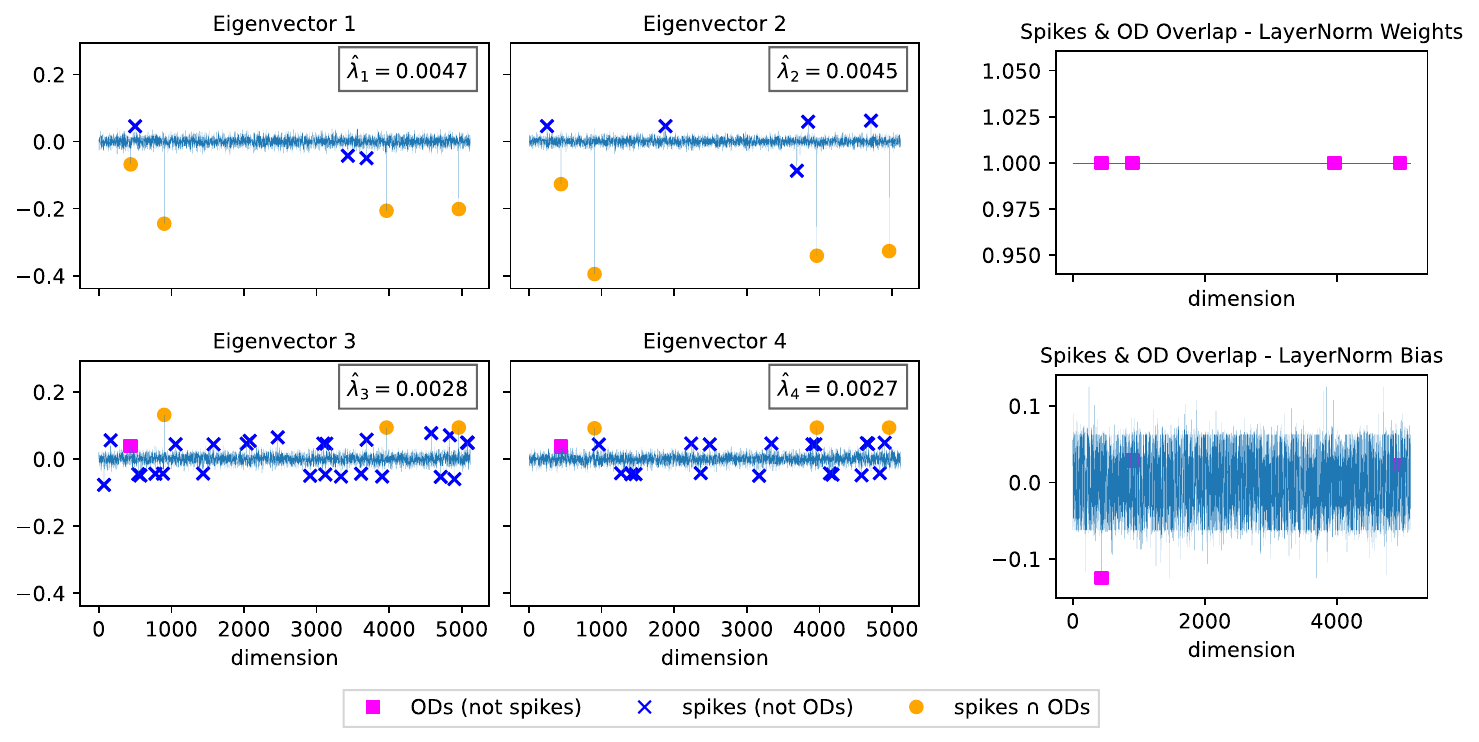}}
\subfloat[gemma-9b]{\includegraphics[scale=0.31]{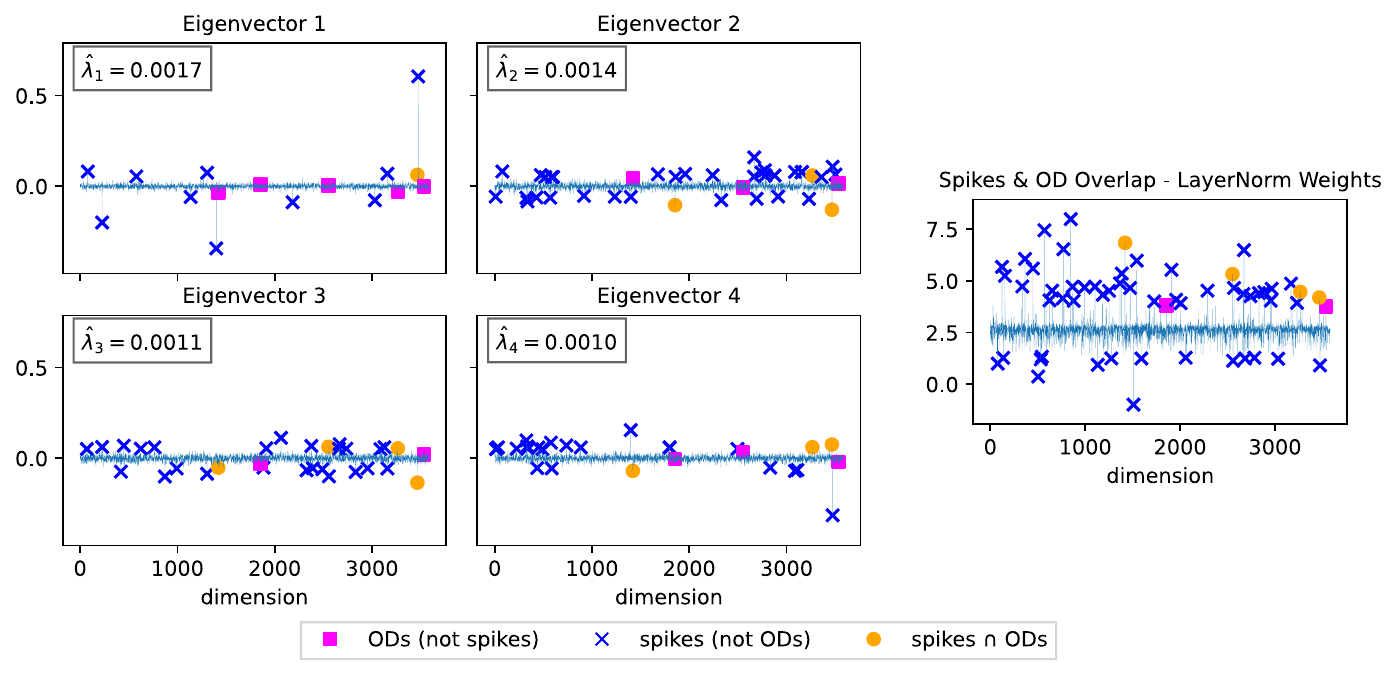}}

\centering
\subfloat[stable-12b]{\includegraphics[scale=0.3]{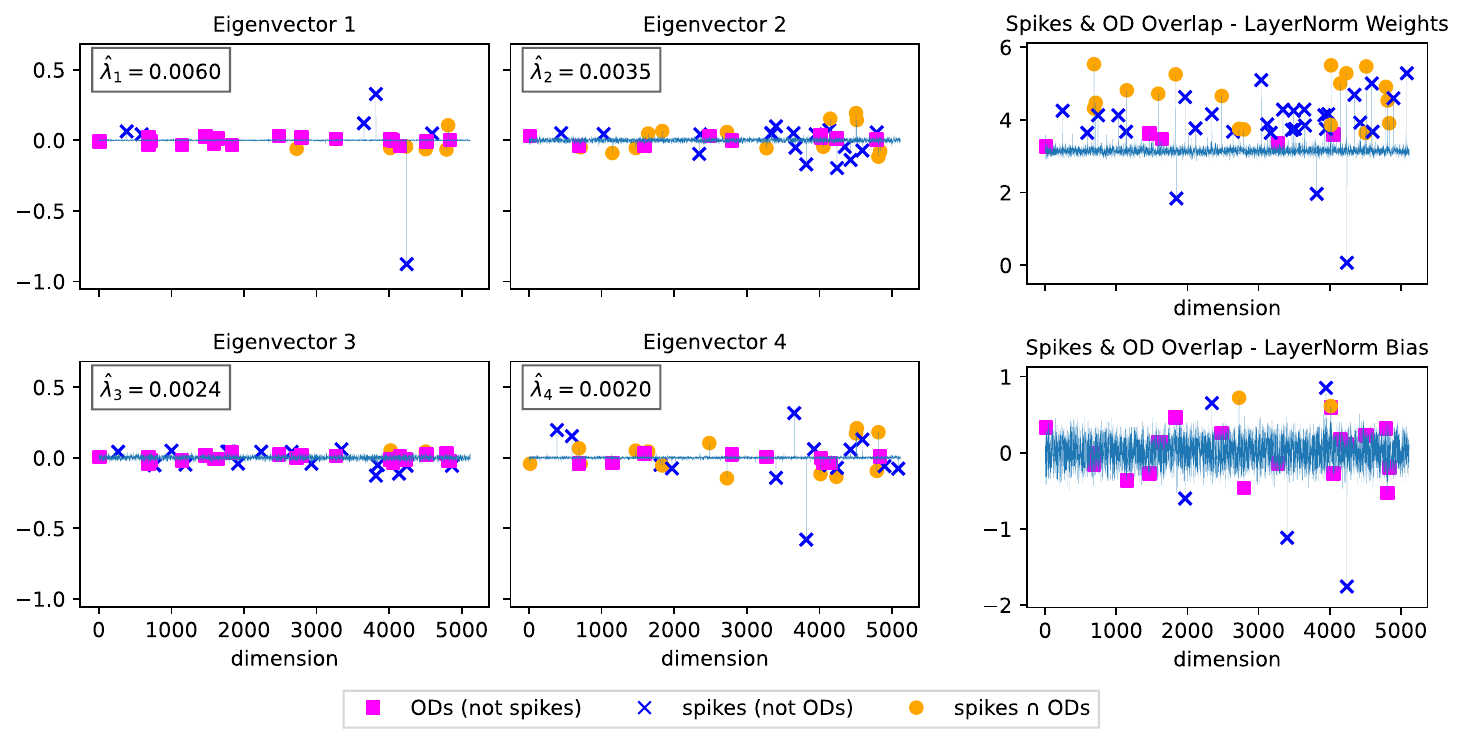}}
\caption{The analysis of Fig.~\ref{fig:pythia-12b_eigenLayerNorm} of the main text is repeated for the other models. \textbf{Left:} Top-4 singular vectors and values of the last-layer MLP down-projection matrix. \textbf{Right:} Final LayerNorm weights and, if present, biases. Spikes in parameter values that correspond to ODs are visualized as orange circles.}\label{fig:OtherModels_eigenLayerNorm}
\end{figure*}

We then select the top 4 singular vectors and repeat the analysis for each of them. Results for pythia-12b are in Fig.~\ref{fig:pythia-12b_eigenLayerNorm} of the main text, and for the other models in Fig.~\ref{fig:OtherModels_eigenLayerNorm} here. The figure shows the overlap between the ODs and the spikes of the singular vectors. It also shows the overlap between the ODs and the (similarly defined) spikes in the last-layer LayerNorm weight and bias vectors.
For all models, we observe a non-negligible overlap of spikes and ODs of their last layer; all of the observed overlaps have \( p \approx 0 \) of occurring by chance according to a random overlap simulation test.

The same analysis applied to the $Q$, $W$, $V$, and $O$ matrices of the attention heads did not reveal any significant overlap between ODs and spike features; therefore, these results are not presented here. While the lack of structure in the $Q$ and $W$ matrices is consistent with their primary function of aggregating information across tokens, we might have expected detectable signals in the $V$ or $O$ matrices. However, no such structure was observed. A possible explanation is that, as ODs act as a fixed bias promoting frequent words, this is more readily encoded in the MLP than in the attention matrices, whose function is to manage contextual information, but we leave further exploration of the difference to future work.


\section{Emergence of ODs during  training}
\label{app:training}

Here, we report the Table~\ref{tab:training_full} characterizing the evolution of ODs across Pythia's checkpoints. We can observe how the accuracy and the role of ODs in predicting frequent tokens evolve during pre-training. As already mentioned in the main text, starting from step 3000 we observe the emergence of a sizable number of ODs, together with an (initially small) negative effect on accuracy upon ablation. At the same time, we see how both the ratios between the number of predicted tokens in a) the only-OD and only-random ablations and b) full model and ablate-OD modalities become smaller. This behavior suggests that the ODs have started specializing in frequent-token prediction.
\begin{table*}[htb]
    \centering
    \setlength{\tabcolsep}{5pt} 
    \begin{tabular}{l|cc|cc|cccc}
    \toprule
    Pythia & \multicolumn{2}{c}{ODs} & \multicolumn{2}{c}{Accuracy [\%]} & \multicolumn{4}{c}{\# Distinct predicted tokens}  \\
    Ckp & \# & $\cap$ final & FM & ablate-OD & FM & ablate-OD & only-OD & only-rnd \\
    \midrule
    500 & 1 & 0 & 11.4 & 11.4 & 667 & 671 & 14 & 20 ± 3 \\
    1000 & 1 & 1 & 17.4 & 17.4 & 2774 & 2790 & 15 & 22 ± 4 \\
    2000 & 7 & 3 & 26.4 & 26.5 & 5168 & 5322 & 63 & 677 ± 130 \\
    3000 & 22 & 7 & 29.6 & 29.4 & 6071 & 6463 & 188 & 2524 ± 649 \\
    4000 & 38 & 11 & 31.1 & 30.7 & 6105 & 6756 & 136 & 4343 ± 1470 \\
    5000 & 42 & 12 & 32.4 & 30.7 & 6472 & 7466 & 157 & 6125 ± 1111 \\
    6000 & 38 & 11 & 33.6 & 31.6 & 6258 & 7361 & 222 & 6541 ± 1245 \\
    7000 & 37 & 10 & 34.1 & 31.6 & 6737 & 7933 & 178 & 6762 ± 673 \\
    8000 & 35 & 9 & 34.4 & 32.1 & 6694 & 8121 & 179 & 6995 ± 475 \\
    10000 & 34 & 11 & 35.5 & 33.2 & 6776 & 8193 & 152 & 5959 ± 563 \\
    12000 & 31 & 10 & 36.1 & 33.9 & 6760 & 8233 & 252 & 4868 ± 750 \\
    14000 & 28 & 10 & 36.3 & 33.5 & 6967 & 8695 & 211 & 5760 ± 715 \\
    16000 & 22 & 9 & 37.6 & 34.3 & 6965 & 8878 & 152 & 5224 ± 246 \\
    64000 & 21 & 18 & 41.1 & 35.7 & 7346 & 10466 & 102 & 5391 ± 139 \\
    143000 & 36 & 36 & 43.0 & 34.3 & 7504 & 11116 & 195 & 7254 ± 269 \\
    \bottomrule
    \end{tabular}
    \caption{ODs presence and behaviour across pythia-12b training checkpoints. \lq{\#}\rq{} indicates the number of ODs, while \lq{$\cap$ final}\rq{} is the number of ODs that are also present at the last checkpoint. The last ckp corresponds to the final model to which we refer in the main text.}
    \label{tab:training_full}
\end{table*}


The only other model for which training checkpoints are public is olmo2-13b. However, the first available checkpoint does not occur early enough, and fully-trained-model OD behavior is already firmly in place by then.
\section{Sentence generation upon ablation}
\label{sec:generation}
In this section we report how the different models' generation capabilities change upon ablation of given dimensions. The prompt is the same for all the models, in order to have a direct comparison. In all cases, we generated 20 tokens with greedy decoding. %
Tokens that could not be properly rendered are substituted in the tables by ``[?]''. These included characters such as U+000B (line tabulation) or U+548C (CJK Unified Ideograph)

Tables \ref{tab:ablation_gen_pythia-12b}-\ref{tab:ablation_gen_stable-12b} show:
\begin{itemize}
    \item default: the sentences generated by the full model 
    \item largest: the sentences generated by ablating the $k$ dimensions with the largest medians
    \item smallest: the sentences generated by ablating the $k$ dimensions with the smallest medians
    \item random: the sentences generated by ablating $k$ random dimensions.
\end{itemize}
In particular, what we call outlier dimensions correspond to the top largest dimension and the effect of their ablation can be found in the first lines of the entries in the ``largest'' block.
We observe that one typically needs to ablate between 2000 and 3000 random or small dimensions for the models to start generating meaningless sentences. Differently, the behavior is much more diversified when ablating the largest dimensions.
Interestingly, pythia-12b stops predicting the token ``a'' (that we identified as OD-favored for this model) after the removal of 5 ODs and the token ``,'' after 20, while repetitions--a clear sign of broken generation--start at 100.
Mistral-7b and llama-8b start producing loops with 10 ablations only and their output becomes meaningless after 50. Qwen-14b, stable-12b and olmo2-13b are the most sensitive, as the ablation of just 5, 15 and 20 dimensions, respectively, are enough to completely disrupt the model generation capabilities.
Opt-13b and gemma-9b are the more robust models, with ablation of the largest dimensions affecting them similarly to the other ablation strategies.

As already stated in the main text, these observations suggest that most models heavily rely on the ODs in order to work properly, to the point that some of them produce meaningless results after the removal of just a few dimensions.  
This also implies that any downstream task involving generation will be affected by the removal of ODs.

\begin{table*}[tb]
\centering\small
\begin{tabular}{r|p{0.9\textwidth}}
Model & pythia-12b \\
\midrule
prompt & Two years ago I learned to bike with my parents. On Sundays' afternoon we always went for a ride. This stopped when my dad broke his knee and needed to go under surgery. All of \\
\midrule
default &  a sudden, I was the only one who could ride. I was so excited to be the only \\
\midrule
Ablation & Generated sentence\\
\midrule
largest 5 &  a sudden, I had to learn how to bike with my mom. She was always nervous about me \\
10 &  us kids were sad. But my mom was sadder. She cried. She cried because she knew \\
15 &  us kids were sad. But my mom said, "You'll learn. You'll learn." And \\
20 &  sudden, cycling became boring.\texttt{\textbackslash n}\texttt{\textbackslash n}Then, when my dad recovered, we went cycling again. \\
30 &  sudden cycling became boring.\texttt{\textbackslash n}\texttt{\textbackslash n}Then came cycling holidays. I've been cycling with my parents since \\
50 &  sudden cycling became boring.\texttt{\textbackslash n}\texttt{\textbackslash n}Then came bike commuting. I've been commuting by bike since February \\
100 &  sudden cycling became boring.\texttt{\textbackslash n}\texttt{\textbackslash n}Then came cycling holidays. cycling holidays are cycling holidays. cycling holidays \\
200 &  sudden cycling became boring. cycling became boring. cycling became boring. cycling became boring. cycling became boring \\
500 &  sudden cycling became boring! :( but yeah! cycling is awesome! :) :) :) :) :) :) :) \\
1000 &  sudden cycling became boring! cycling became boring! cycling became boring! cycling became boring! cycling became boring \\
2000 &  sudden cycling wasn't fun anymore anymore anymore anymore anymore anymore anymore anymore anymore anymore anymore anymore anymore anymore anymore \\
3000 &  sudden cycling seemed boring! Cycling routes arent longer anymore! Cycling routes arent longer anymore \\
4000 &  sudden cycling wasn't FUNNIERORANDORANDORANDORANDORANDORANDORANDORANDORANDORANDORANDORANDORAND \\
5000 & implementation GU Dimelsleroil Hayes Weaverhet LibrarydocumentclassenvabcDEF[?]atchningriumt \\
\midrule
smallest 5 &  a sudden, I was the only one who could ride. I was so excited to be the only \\
10 &  a sudden, I had to take care of my mom and my brother. I was a little bit \\
15 &  a sudden, I had to take care of my mom and my brother. I had to take care \\
20 &  a sudden I was the only one who could ride with my parents. I was so excited to be \\
30 &  a sudden, I had to take care of my mom and my brother. I was the only one \\
50 &  a sudden, I had to take care of my mom and my brother. I had to take care \\
100 &  a sudden I had to learn to bike on my own. I was scared and nervous. I was \\
200 &  a sudden I had to learn to bike on my own. I was scared, but I did it \\
500 &  a sudden I had to take care of my mom and my brother. I had to take care of \\
1000 &  a sudden, I had to take the bike on my own. I was not a good biker \\
2000 &  the sudden the biking was a thing of the past, and the only thing that was still the \\
3000 &  the biking in the world, the two of the three bikely-sib-b \\
4000 &  the first bikely enthusic-tic enthusic enthusic enthusic enthusic enthusic \\
5000 &  the the the the the the the the the the the the the the the the the the the the \\
\midrule
random 5 &  a sudden, I was the only one who could ride. I was so excited to be the only \\
10 &  a sudden, I was the only one who could ride. I was so excited to be the only \\
15 &  a sudden, I was the only one who could ride. I was so excited to be the only \\
20 &  a sudden, I was the only one who could ride. I was so excited to be the only \\
30 &  a sudden, I had to take care of my mom and my brother. I was the only one \\
50 &  a sudden, I was the only one who could ride. I was so excited to be the only \\
100 &  a sudden I was the only one who could ride with my parents. I was so excited to be \\
200 &  a sudden I had to take care of my mom and my brother. I was a little bit scared \\
500 &  a sudden I had to take care of my mom and my brother. I had to do everything. \\
1000 &  us kids were left with our bikes in the garage. My mom asked me to take them out for \\
2000 &  us kids wanted to help him but his answer was always the same [?] "I can manage". This \\
3000 &  these long rides then became very short and very very very very very very very very very very very very \\
4000 &  us converged over distance over distance over distance over distance over distance over distance over distance over distance over \\
5000 & typentypen photograp tiss scrut infil ingredtypentypentypentypentypentypentypentypentypentypentypentypentypen \\
\bottomrule
\end{tabular}
\caption{Generation upon ablation for pythia-12b. See Appendix \ref{sec:generation} for details.}
\label{tab:ablation_gen_pythia-12b}
\end{table*}
\begin{table*}[tb]
\centering\small
\begin{tabular}{r|p{0.9\textwidth}}
Model & mistral-7b \\
\midrule
prompt & Two years ago I learned to bike with my parents. On Sundays' afternoon we always went for a ride. This stopped when my dad broke his knee and needed to go under surgery. All of \\
\midrule
default &  a sudden I was alone. I didn't know how to bike without my parents. I was \\
\midrule
Ablation & Generated sentence\\
\midrule
largest 5 &  a sudden I was alone.\texttt{\textbackslash n}\texttt{\textbackslash n}I've always been a bit of a loner. \\
10 &  a sudden, I was alone.\texttt{\textbackslash n}\texttt{\textbackslash n}I've been riding my bike for 2 years \\
15 &  a sudden, I was alone.\texttt{\textbackslash n}\texttt{\textbackslash n}  I've been riding my bike for 2 \\
20 &  a sudden, I was alone.\texttt{\textbackslash n}\texttt{\textbackslash n}  I've been riding alone for 2 years \\
30 &  a sudden, I was alone.\texttt{\textbackslash n}\texttt{\textbackslash n}I'm 16, and I'm \\
50 &  a sudden, I'm 16, and I'm alone.\texttt{\textbackslash n}\texttt{\textbackslash n}I' \\
100 &  a sudden, I'm 18, and I'm 180cm tall \\
200 &  a sudden, I'm 16, and I'm biking.\texttt{\textbackslash n}\texttt{\textbackslash n}Geplaatst \\
500 &  a sudden, biking with my parents, and specially, with my dad, 10 \\
1000 &  us, including mum, are avid cyclists. Mum's been cycling since she' \\
2000 &  a sudden, biking with him, which I've grown accustomed, I've \\
3000 &  Suddenly bike rides turned into walks. Weekends turned into sitting. Weekends turned into \\
4000 &  [[?]] [[?]] [[?]] [[?]] [[?]] [[?]] [[?]] [[?]] [[?]] [[?]] [[?]] [[?]] [[?]] [[?]] [[?]] [[?]] [[?]] [[?]] [[?]] [[?]] \\
\midrule
smallest 5 &  a sudden I was alone. I didn't know how to bike without my parents. I was \\
10 &  a sudden I was alone. I didn't know how to bike without my parents. I was \\
15 &  a sudden I was alone. I didn't know how to bike without my parents. I was \\
20 &  a sudden I was alone. I didn't know how to bike without my parents. I was \\
30 &  a sudden I was alone. I didn't know how to bike without my parents. I was \\
50 &  a sudden I was alone. I had to find a new hobby.\texttt{\textbackslash n}\texttt{\textbackslash n}I started to \\
100 &  a sudden I was alone. I had to find a new hobby.\texttt{\textbackslash n}\texttt{\textbackslash n}I started to \\
200 &  a sudden I was alone. I had to learn to bike by myself. I was scared. I \\
500 &  a sudden I had to bike on my own. I was scared to go on my own, so \\
1000 &  a sudden I had to learn to ride by myself. I started with the bike my parents had given \\
2000 &  this time I had a little dream to go to the other side of the world and travel by bike \\
3000 &  this caused my love for the mountain bike to start slowly slowly slowly slowly slowly slowly slowly slowly slowly slowly \\
4000 &  laugtypenplaattypentypentypentypentypentypentypentypentypentypentypentypentypentypentypentypentypen \\
\midrule
random 5 &  a sudden I was alone. I had to find a new hobby.\texttt{\textbackslash n}\texttt{\textbackslash n}I started to \\
10 &  a sudden I was alone. I didn't know how to bike without my parents. I was \\
15 &  a sudden I was alone. I didn't know how to bike without my parents. I was \\
20 &  a sudden I was alone with my bike. I didn't know how to ride it. I \\
30 &  a sudden I was alone with my bike. I didn't know how to ride it. I \\
50 &  a sudden I was alone. I didn't know how to bike without my parents. I was \\
100 &  a sudden I was alone. I didn't know how to bike without my parents. I was \\
200 &  a sudden I was alone with my bike. I didn't know how to ride it anymore. \\
500 &  a sudden I was alone with my bike and didn't know what to do with it. I \\
1000 &  a sudden I was on my own and I didn't know how to ride a bike anymore. \\
2000 &  a sudden my mom and I were on our own and we really enjoyed it!\texttt{\textbackslash n}\texttt{\textbackslash n}A few \\
3000 &  those things mean that we can do that in these hard corona-19 (C-1 \\
4000 & friedunoweenesen interesting tempornen interesting temporopleunstöß dedößößößößößößöß \\
\bottomrule
\end{tabular}
\caption{Generation upon ablation for mistral-7b. See Appendix \ref{sec:generation} for details.}
\label{tab:ablation_gen_mistral-7b}
\end{table*}
\begin{table*}[tb]
\centering\small
\begin{tabular}{r|p{0.9\textwidth}}
Model & llama-8b \\
\midrule
prompt & Two years ago I learned to bike with my parents. On Sundays' afternoon we always went for a ride. This stopped when my dad broke his knee and needed to go under surgery. All of \\
\midrule
default &  a sudden I was left alone with my bike. I was 12 years old and I had to \\
\midrule
Ablation & Generated sentence\\
\midrule
largest 5 &  this happened during summer. Now that winter is coming I'm afraid that biking will stop again. I \\
10 &  this happened during summer 2012. Since then I haven't ridden my bike anymore. Until yesterday \\
15 &  us were devastated. I couldn't believe it. I couldn't believe my dad couldn't bike anymore \\
20 &  a sudden, biking became dangerous. I've been biking alone for a year now.<|end\_of\_text|> \\
30 &  a sudden, biking became dangerous. I've been biking alone 2-3 times a week. \\
50 &  a sudden, biking became dangerous. I've been biking alone 2-3 times a week. \\
100 &  2 months 2.scalablytyped[?][?][?][?] full[?][?][?][?][?][?] \\
200 &  2 summers' fkk fkk fkk fkk fkk fkk fkk fkk fkk fkk fkk fkk fkk fkk fkk fkk \\
500 &  2 weeks 2.scalablytyped nettsteder datingside datingside datingside datingside datingside datingside datingside datingside datingside datingside datingside datingside datingside \\
1000 &  2 summers ago 2 summers 2 summers 2 summers 2 summers 2 summers  \\
2000 &  2012  passed me biking dreaming  'gc addCriterionHeaderCodeHeaderCodeHeaderCodeHeaderCodeHeaderCodeHeaderCodeHeaderCode \\
3000 &  2012 2013 2014 2015 [?]\#[?]\#[?]\#[?]\#[?]\#[?]\#[?]\# \\
4000 &  thisuskollectors@natenbspnbspgroupongroupongroupongroupongroupongroupongroupongroupon[?][?][?][?]\\
\midrule
smallest 5 &  a sudden I was left alone with my bike. I was 12 years old and I had to \\
10 &  a sudden I was left alone with my bike. I was not able to ride it anymore. I \\
15 &  a sudden I was left alone with my bike. I was 12 years old and I was not \\
20 &  a sudden I was left alone with my bike. I was not able to ride it anymore. I \\
30 &  a sudden I was left alone with my bike. I was not sure if I could ride it by \\
50 &  a sudden I was left alone with my bike. I was not able to ride it anymore. I \\
100 &  a sudden I was left alone with my bike. I was 12 years old, and I was \\
200 &  a sudden I was left alone with my bike. I was 12 years old, I had a \\
500 &  a sudden I was the only one who could ride a bike. I was 11 years old. \\
1000 &  a sudden I, a 15 year old,  was in charge of the  a \\
2000 &  these, 730 830   1 830<|end\_of\_text|> \\
3000 &  enthusi         usercontent/company-        \\
4000 & HeaderCode                    \\
\midrule
random 5 &  a sudden I was left alone with my bike. I was not able to ride it anymore. I \\
10 &  a sudden I was left alone with my bike. I was not able to go for a ride with \\
15 &  a sudden I was left alone with my bike. I was not able to ride it anymore. I \\
20 &  a sudden I was left alone with my bike. I was not able to ride it anymore. I \\
30 &  a sudden I was left alone with my bike. I was not able to go for a ride with \\
50 &  a sudden I was left alone with my bike. I was sad, but I didn't want to \\
100 &  a sudden I was left alone with my bike. I started to ride around the block, then around \\
200 &  a sudden I was alone and I didn't know how to ride a bike. I was scared to \\
500 &  a sudden I was left alone with my bike and I was sad. I missed my dad and I \\
1000 &  a sudden I was left with no one to ride with and no one to ride for.\texttt{\textbackslash n}I miss \\
2000 &  a sudden we had no reason to meet and ride in the fields and fields of the neighbouring area where \\
3000 &  a sudden the non use of a bike just for ñana  or ñana  or \\
4000 & \_marvin éné  //\{\{1\_//\{\{ éné iste  //\{\{  //\{\{ \\
\bottomrule
\end{tabular}
\caption{Generation upon ablation for llama-8b. See Appendix \ref{sec:generation} for details.}
\label{tab:ablation_gen_llama-8b}
\end{table*}
\begin{table*}[tb]
\centering\small
\begin{tabular}{r|p{0.9\textwidth}}
Model & olmo2-13b \\
\midrule
prompt & Two years ago I learned to bike with my parents. On Sundays' afternoon we always went for a ride. This stopped when my dad broke his knee and needed to go under surgery. All of \\
\midrule
default &  a sudden I was too busy to go biking. I was always doing something else. My parents asked \\
\midrule
Ablation & Generated sentence\\
\midrule
largest 5 &  us were sad about it. Yesterday I went to my dad's room. He was lying on the \\
10 &  us were sad about it. Yesterday, my dad told me that he wanted to ride bikes again. \\
15 &  us were sad about it. Yesterday morning mom told me dad's knee was healed and he could ride \\
20 &  us were sad about it.[?][?][?][?][?][?][?][?][?][?][?][?][?][?] \\
30 &  us were sad about it.[: \_ Z \_ X \_ X \_ K][?][?][?][?] \\
50 &  sudden[?][?][?][?][?][?][?][?][?][?][?][?][?][?][?][?][?][?][?] \\
100 &  sudden[?][?]\texttt{\textbackslash n}\texttt{\textbackslash n}KANJI[?][?][?]\texttt{\textbackslash n}\texttt{\textbackslash n}KANJI[?][?][?]\texttt{\textbackslash n}\texttt{\textbackslash n}KANJI[?][?][?]\texttt{\textbackslash n}\texttt{\textbackslash n}KANJI[?][?]\\
200 &  sudden[?][?]\texttt{\textbackslash n}\texttt{\textbackslash n}KANJI[?][?][?][?][?][?][?][?][?][?][?][?]\\
500 &  sudden[?][?]\texttt{\textbackslash n}\texttt{\textbackslash n}KANJI[?][?][?][?][?][?][?][?][?][?][?][?]\\
1000 &  sudden[?][?]\texttt{\textbackslash n}\texttt{\textbackslash n}KANJI[?][?][?][?][?][?][?][?][?][?][?][?]\\
2000 &  sudden biking Sundays Stops[?][?]\texttt{\textbackslash n}\texttt{\textbackslash n}KANJI[?][?][?]\texttt{\textbackslash n}\texttt{\textbackslash n}KANJI[?][?][?]\texttt{\textbackslash n}\texttt{\textbackslash n}KANJI[?][?][?]\texttt{\textbackslash n}\texttt{\textbackslash n}KANJI[?][?][?]\\
3000 &  sudden(weather[?][?][?][?][?][?][?][?][?][?][?][?][?][?][?][?]\\
4000 &  Eagerness accordionistarween accordistarween accordistarween accordistarween accordistarween accordistarween \\
5000 & ayasundo Lyons EOamelbitsujuenkinkeenk KramerFParnessVTesch Estragos backlogeriicode \\
\midrule
smallest 5 &  a sudden I was too busy to go biking. I was always doing something else. My parents asked \\
10 &  a sudden I was too busy to go biking. I was always doing something else. My parents would \\
15 &  a sudden I was too busy to go biking. I was always doing something else. My parents would \\
20 &  a sudden I was too busy to go biking. I was always doing something else. My parents would \\
30 &  a sudden I was too busy to go biking. I was always doing something else. My dad was \\
50 &  a sudden I was too busy to go biking. I was always doing something else. My dad was \\
100 &  a sudden I was too busy to go biking. I was always doing something else. I had forgotten \\
200 &  a sudden my mum was too tired to go out, and I didn't have a bike for myself \\
500 &  a sudden my mum was too tired to go out and I was too lazy to go alone. So \\
1000 &  a sudden I was the one who had to take the car to the store to get the milk and \\
2000 &  the other times we had to go to the hospital to get a new treatment. This was a very \\
3000 &  the other days I would go to the park and go on the little dirt road and go to the \\
4000 &  the  15  10  5  1  1  1  \\
5000 &  the  100- 100-  100-  100  100   \\
\midrule
random 5 &  a sudden I was too busy to go biking. I was always doing something else. My parents asked \\
10 &  a sudden I was too busy to go biking. I was always doing something else. My parents asked \\
15 &  a sudden I was too busy to go biking. I was always doing something else. My parents asked \\
20 &  a sudden I was too busy to go biking. I was always doing something else. My parents asked \\
30 &  a sudden I was too busy to go biking. I was always doing something else. My parents asked \\
50 &  a sudden I was too busy to go biking. I was always doing something else. I had a \\
100 &  a sudden I was too busy to go biking. I was always doing something else. I had a \\
200 &  a sudden, my mum and I stopped riding. We were too afraid to ride without him. One \\
500 &  a sudden, my mom was the only one who could bike with me. But she was very busy \\
1000 &  a sudden my mom was the only one who could bike with me. My dad was always the one \\
2000 &  a sudden, my mom was the only one who biked. I tried to get her to do \\
3000 &  sudden I found myself riding the most I ever rode before. Mom continued to drive her car and Dad \\
4000 &  sudden mid-November central Europe weather hit home. Snow Brussels Wallonia  year old electric bike  \\
5000 & tahuncente[?] presenta[?]órioPerfilVocêArchivoVocêVocêVocêVocêVocêVocêVocêVocêVocêVocêVocê \\
\bottomrule
\end{tabular}
\caption{Generation upon ablation for olmo2-13b. See Appendix \ref{sec:generation} for details.}
\label{tab:ablation_gen_olmo2-13b}
\end{table*}
\begin{table*}[tb]
\centering\small
\begin{tabular}{r|p{0.9\textwidth}}
Model & qwen-14b \\
\midrule
prompt & Two years ago I learned to bike with my parents. On Sundays' afternoon we always went for a ride. This stopped when my dad broke his knee and needed to go under surgery. All of \\
\midrule
default &  a sudden I was left without a biking partner. I was sad and I missed the rides. I \\
\midrule
Ablation & Generated sentence\\
\midrule
largest 5 &  my[?]200[?]bike[?]lessons[?]were[?]wasted.[?]I[?]knew \\
10 &  us volunteerd to volunte yarg yarg yarg yarg yarg yarg yarg yarg yarg yarg yarg yarg yarg yarg yarg \\
15 &  2012 I biked 12000000000 \\
20 &  2012121212121212121 \\
30 &  2012 1 [?] be [?]spent in[?][?] [?][?][?][?] and he hasn \\
50 &  2012 1 [?] be [?]spent in[?][?] [?][?][?][?] and then[?] \\
100 &  2012 1)didn't [?])did[?] [?])did[?] [?])did[?] [?] \\
200 &  2012 1)didn't [?])did[?] [?])did[?] [?])did[?] [?] \\
500 &  2012 1)didn't [?])did[?] [?])did[?] [?])did[?] [?] \\
1000 &  2012 1)did [?])did[?] [?])did[?] [?])did[?] [?])did[?] \\
2000 &  2012, he hasn't[?][?][?][?][?][?][?][?][?][?][?][?][?][?][?][?][?][?][?][?][?][?] \\
3000 &  2015,[?][?][?][?][?][?][?][?][?][?][?][?][?][?][?][?][?][?][?][?][?][?][?][?][?] \\
4000 & us\$core\$core\$core\$core\$core\$core\$core\$core\$core\$core\$core\$core\$core\$core\$core\$core\$core\$core\$core \\
5000 & [?]s own behalf destinationViewControlleruallyALLY [?][?]\texttt{\textbackslash n}[?] [?][?]ses [?][?]ses [?][?]ses [?][?]ses [?][?]ses \\
\midrule
smallest 5 &  a sudden I was left without a biking partner. I was sad and I missed the rides. I \\
10 &  a sudden I was left without a biking partner. I was sad and I missed the rides. I \\
15 &  a sudden I was left without a biking partner. I was sad and I missed the rides. I \\
20 &  a sudden I was left without a biking partner. I was sad and I missed the rides. I \\
30 &  a sudden I was without a bike and without a dad to ride with. I was sad. I \\
50 &  a sudden I was without a bike. I was sad. I was 10 years old. \\
100 &  a sudden I had to stop biking. I was sad. I missed biking. I missed the wind \\
200 &  a sudden I had to stop biking. I was sad. I missed the wind in my face, \\
500 &  a sudden I had to stop riding. I felt so sad. I had to stay home and watch \\
1000 &  a sudden I had no one to ride with. I had a hard time to get used to the \\
2000 &  his  post-surgery  care was very hard on him and he never got back to biking. \\
3000 &  his  left  [?]  [?][?][?][?][?][?][?][?][?][?][?][?] \\
4000 &  juni  I  icked      1  ets  j   \\
5000 & FunctionFlags  [?][?][?][?][?][?]               \\
\midrule
random 5 &  a sudden I was left without a biking partner. I was sad and I missed the rides. I \\
10 &  a sudden I was left without a biking partner. I was sad and I missed the rides. I \\
15 &  a sudden I was left without a biking partner. I was sad and I missed the time we spent \\
20 &  a sudden I was left without a biking partner. I was sad and I didn't know what to \\
30 &  a sudden I was left without a bike. I was 10 years old and I didn't \\
50 &  a sudden I was left without a bike. I was 12 years old and I didn't \\
100 &  a sudden I was left without a bike. I was 10 years old and I didn't \\
200 &  a sudden my mom and I were left alone. I was 10 years old and my mom \\
500 &  a sudden my 10-year-old self was the only one left to take care of my little \\
1000 &  sudden we had to stop going to bike together because my dad could not stand up yet to go for \\
2000 &  sudden dad could not go cycling anymore but he was not going to  \_ his 2-year-old \\
3000 &  newcoms([?][?])[?][?]f111111111111 \\
4000 & [?]ALSE [?][?][?][?] [?][?][?][?]\$fdata\$fdata[?]\$fdata[?]\$fdata\$fdata\$fdata\$fdata\$fdata\$fdata\$fdata\$fdata\$fdata\$fdata\$fdata \\
5000 & .dtpcola-kit/Branch/goto/portfolio/portfolio, volunte/lic/gif/Branch/goto/portfolio/portfolio/portfolio, volunte/chart/Branch \\
\bottomrule
\end{tabular}
\caption{Generation upon ablation for qwen-14b. See Appendix \ref{sec:generation} for details.}
\label{tab:ablation_gen_qwen-14b}
\end{table*}
\begin{table*}[tb]
\centering\small
\begin{tabular}{r|p{0.9\textwidth}}
Model & opt-13b \\
\midrule
prompt & Two years ago I learned to bike with my parents. On Sundays' afternoon we always went for a ride. This stopped when my dad broke his knee and needed to go under surgery. All of \\
\midrule
default &  a sudden I was alone on the bike. I was scared to ride alone. I was scared to \\
\midrule
Ablation & Generated sentence\\
\midrule
largest 5 &  a sudden I was alone on the bike. I was scared and didn't know what to do. \\
10 &  a sudden I was alone. I was so sad. I miss my dad.\texttt{\textbackslash n}I'm sorry \\
15 &  a sudden I was alone. I was so sad. I didn't know what to do. I \\
20 &  a sudden I was alone. I was afraid to ride my bike alone. I was afraid to ride \\
30 &  a sudden I was alone. I was so sad. I didn't know what to do. I \\
50 &  a sudden I was alone. I was afraid to ride my bike. I was afraid to go outside \\
100 &  a sudden I was alone. I was so sad. I miss my dad.\texttt{\textbackslash n}I'm sorry \\
200 &  a sudden I was alone. I was so sad. I was so lonely. I was so angry \\
500 &  a sudden I was the only one who could go for a ride. I was so sad. I \\
1000 &  us are getting older and we don't have the energy anymore. I'm really happy I'm getting \\
2000 &  us are getting older. My mom doesn't want to ride anymore. My dad doesn't want to \\
3000 &  us are cyclists. Cycling is our favorite thing. Cycling is our favorite thing. Cycling is our favorite \\
4000 &  my siblings bike, I'm alone. :(\texttt{\textbackslash n}bike on!</s> \\
5000 &  Oxacander PVvblogerherlabeleer missvenalsiche Jeriter Monitoring materialsrugeer \\
\midrule
smallest 5 &  a sudden I was alone on the bike. I was scared to ride alone. I was scared to \\
10 &  a sudden I was alone on the bike. I was scared to ride alone. I was scared to \\
15 &  a sudden I was alone on the bike. I was scared and I didn't know what to do \\
20 &  a sudden I was alone on the bike. I was scared and I didn't know what to do \\
30 &  a sudden I was alone on the bike. I was scared and I didn't know what to do \\
50 &  a sudden I was alone. I was so sad. I was so lonely. I was so scared \\
100 &  a sudden I was alone on the bike. I was scared and I didn't know what to do \\
200 &  a sudden I was the only one who could ride a bike. I was so sad.\texttt{\textbackslash n}\texttt{\textbackslash n} \\
500 &  a sudden I was the only one left in the family who could ride a bike. I was the \\
1000 &  a sudden I was the only one left in the family who could ride a bike.\texttt{\textbackslash n}\texttt{\textbackslash n}I \\
2000 &  a sudden the time for the bike rides was over.\texttt{\textbackslash n}\texttt{\textbackslash n}I was a bit disappointed, but \\
3000 &  a sudden , the only time , , , , , , , , , , , , , , \\
4000 &  , , , , , , , , , , , , , , , , , , , , \\
5000 &  .\_ . . . . . . . . . . . . . . . . . . . \\
\midrule
random 5 &  a sudden I was alone on the bike. I was scared and I didn't know what to do \\
10 &  a sudden I was alone on the bike. I was scared to ride alone. I was scared to \\
15 &  a sudden I was alone on the bike. I was scared to ride alone. I was scared to \\
20 &  a sudden I was alone on the bike. I was scared and I didn't know what to do \\
30 &  a sudden I was alone on the bike. I was scared and I didn't know what to do \\
50 &  a sudden I was alone on the bike. I was not able to ride with my parents anymore. \\
100 &  a sudden I was alone on the bike. I was not able to ride with my parents anymore. \\
200 &  a sudden I was alone on the bike. I was so scared that I stopped riding. I was \\
500 &  a sudden I was alone. I was so sad. I was so lonely. I was so bored \\
1000 &  a sudden I was alone. I was so happy when my mom said that she would take me on \\
2000 &  us are getting older and I can't really ride with my mom and my sister. I miss my \\
3000 &  us missed it and I missed it the most. I miss my dad and I miss my mom who \\
4000 &  a sudden, my mom got lost on her scooter on a daily trip between shops, and lost \\
5000 &  Course permanentalon Pug pendataneptri Irugi pendatanrientht Allegippy Pug pendatanrient \\
\bottomrule
\end{tabular}
\caption{Generation upon ablation for opt-13b. See Appendix \ref{sec:generation} for details.}
\label{tab:ablation_gen_opt-13b}
\end{table*}
\begin{table*}[tb]
\centering\small
\begin{tabular}{r|p{0.9\textwidth}}
Model & gemma-9b \\
\midrule
prompt & Two years ago I learned to bike with my parents. On Sundays' afternoon we always went for a ride. This stopped when my dad broke his knee and needed to go under surgery. All of \\
\midrule
default &  a sudden I was left alone with my bike. I was 10 years old and I was \\
\midrule
Ablation & Generated sentence\\
\midrule
largest 5 &  a sudden I was left alone with my bike. I was afraid to go out alone, but I \\
10 &  a sudden I was left alone with my bike. I was afraid to ride it alone, but I \\
15 &  a sudden I was left alone with my bike. I was scared to ride it alone, but I \\
20 &  a sudden I was left alone with my bike. I was afraid to ride it alone, but I \\
30 &  a sudden I was left alone with my bike. I was afraid to ride it alone, but I \\
50 &  a SUDDEN I was left alone with my bike. I was scared to ride it alone. I \\
100 &  us were sad about it.\texttt{\textbackslash n}\texttt{\textbackslash n}I was sad because I loved biking with my parents. I was \\
200 &  us were sad about it. My mom and I were sad because we couldn XNUMXt ride anymore \\
500 &  us were sad about it. My dad was sad because he couldn XNUMXt ride his bike anymore \\
1000 &  us were devastated. My dad was always so active and sporty. He loved to play tennis, soccer \\
2000 &  us were sad about it. My dad was sadder than anyone else. He loves biking. He \\
3000 &  this stopped when I went to Beijing.\texttt{\textbackslash n}\texttt{\textbackslash n} plea[?]ure\texttt{\textbackslash n} createSlice\texttt{\textbackslash n} I\texttt{\textbackslash n} I\texttt{\textbackslash n} I\texttt{\textbackslash n} I \\
\midrule
smallest 5 &  a sudden I was left alone with my bike. I was 10 years old and I was \\
10 &  a sudden I was left alone with my bike. I was scared to go out alone, but I \\
15 &  a sudden I was left alone with my bike. I was scared to go out alone, but I \\
20 &  a sudden I was left alone with my bike. I was scared to go out alone, but I \\
30 &  a sudden I was left alone with my bike. I was 12 years old and I was \\
50 &  a sudden I was left alone with my bike. I was 12 years old and I was \\
100 &  a sudden I was left alone with my bike. I was 12 years old and I was \\
200 &  a sudden I was left alone with my bike. I was 12 years old and I was \\
500 &  a <strong>sudden</strong> I was left alone with my bike. I was 12 years \\
1000 &  a <strong>sudden</strong>, I was left alone. I was 12 years <strong> \\
2000 &  a 2019 I was in a big and very very very very very very very very \\
3000 &  the 2019 and 2020, I was in the 10 \\
\midrule
random 5 &  a sudden I was left alone with my bike. I was 10 years old and I was \\
10 &  a sudden I was left alone with my bike. I was 10 years old and I was \\
15 &  a sudden I was left alone with my bike. I was 10 years old and I was \\
20 &  a sudden I was left alone with my bike. I was 10 years old and I was \\
30 &  a sudden I was left alone with my bike. I was 10 years old and I was \\
50 &  a sudden I was left alone with my bike. I was 10 years old and I was \\
100 &  a sudden I was left alone with my bike. I was 10 years old and I was \\
200 &  a sudden I was left alone with my bike. I was not sure what to do with it. \\
500 &  a sudden I was left alone with my bike. I was afraid to ride it alone, but I \\
1000 &  a sudden I was left alone with my bike. I was afraid to go out alone, but I \\
2000 &  a sudden I was on my own. I was afraid to ride alone, but I knew I had \\
3000 &  a extranjero[?]jimo. ]=> ]=> ]=> ]=> ]=> ]=> ]=> ]=> \\
\bottomrule
\end{tabular}
\caption{Generation upon ablation for gemma-9b. See Appendix \ref{sec:generation} for details.}
\label{tab:ablation_gen_gemma-9b}
\end{table*}
\begin{table*}[tb]
\centering\small
\begin{tabular}{r|p{0.9\textwidth}}
Model & stable-12b \\
\midrule
prompt & Two years ago I learned to bike with my parents. On Sundays' afternoon we always went for a ride. This stopped when my dad broke his knee and needed to go under surgery. All of \\
\midrule
default &  us were very sad. I was the only one who could still ride a bike. I was very \\
\midrule
Ablation & Generated sentence\\
\midrule
largest 5 &  a sudden, I was the only one who could go for a ride. I was 11 \\
10 &  a sudden, I was the only one who could go for a ride. I was 11 \\
15 &  a\texttt{\textbackslash n}s\texttt{\textbackslash n}s\texttt{\textbackslash n}s\texttt{\textbackslash n}s\texttt{\textbackslash n}s\texttt{\textbackslash n}s\texttt{\textbackslash n}s\texttt{\textbackslash n}s\texttt{\textbackslash n}s\texttt{\textbackslash n} \\
20 &  a 110 111 112 113 11 \\
30 &  a 111111111111111111 \\
50 &  a 110 111 112 113 11 \\
100 &  a 110 111 112 113 11 \\
200 &  a\texttt{\textbackslash n}\_.\texttt{\textbackslash n}\_2\_2\_2\_2\_2\_2\_2\_2 \\
500 & \_0\_2\_2\_2\_2\_2\_2\_2\_2\_2 \\
1000 & \_1\_\_2\_3\_4\_5\_6\_7\_8\_9\_ \\
2000 &  dit dit dit dit dit dit dit dit dit dit dit dit dit dit dit dit dit dit dit dit \\
3000 &  moo moo moo moo moo moo moo moo moo moo moo moo moo moo moo moo moo moo moo moo \\
4000 &  sic sic sic sic sic sic sic sic sic sic sic sic sic sic sic sic sic sic sic sic \\
5000 & YELLOW ess repentDomain rit sic wel pens wel wel wel wel wel wel wel wel wel wel wel wel \\
\midrule
smallest 5 &  us were very sad. I was the only one who could still ride a bike. I was very \\
10 &  us were very sad. I was the only one who could still ride a bike. I was very \\
15 &  us were very sad. I was the only one who could still ride a bike. I was very \\
20 &  us were very sad. I was the only one who could still ride a bike. I was very \\
30 &  us were very sad. I was the only one who could still ride a bike. I was very \\
50 &  us were very sad. I was the only one who could still ride a bike. I was very \\
100 &  us were very sad. I was the only one who could still ride a bike. I was very \\
200 &  us were very sad. We missed the Sunday bike rides. We missed the time we spent together. \\
500 &  us were very sad. I was the only one who could still go out on the road. I \\
1000 &  a sudden, I was the only one in the family who could still go for a ride. I \\
2000 &  excer excer excer excer excer excer excer excer excer excer excer excer excer excer excer excer excer excer excer excer \\
3000 &  disappe,[?]\texttt{\textbackslash n}\texttt{\textbackslash n},[?]\texttt{\textbackslash n}\texttt{\textbackslash n},[?]\texttt{\textbackslash n}\texttt{\textbackslash n},[?]\texttt{\textbackslash n}\texttt{\textbackslash n},[?]\texttt{\textbackslash n}\texttt{\textbackslash n},[?]\texttt{\textbackslash n}\texttt{\textbackslash n},[?]\texttt{\textbackslash n}\texttt{\textbackslash n},[?]\texttt{\textbackslash n}\texttt{\textbackslash n},[?]\texttt{\textbackslash n}\texttt{\textbackslash n},[?]\texttt{\textbackslash n}\texttt{\textbackslash n},[?]\texttt{\textbackslash n}\texttt{\textbackslash n},[?]\texttt{\textbackslash n}\texttt{\textbackslash n},
\\
4000 & ,[?]\texttt{\textbackslash n}\texttt{\textbackslash n}INCLUDED,[?]\texttt{\textbackslash n}\texttt{\textbackslash n}INCLUDED,[?]\texttt{\textbackslash n}\texttt{\textbackslash n}INCLUDED,[?]\texttt{\textbackslash n}\texttt{\textbackslash n}INCLUDED,[?]\texttt{\textbackslash n}\texttt{\textbackslash n}INCLUDED,
\\
5000 &  underminhsiLOCKSLOCKSLOCKSLOCKSLOCKSLOCKSLOCKSLOCKSLOCKSLOCKSLOCKS
\\
\midrule
random 5 &  us were very sad. I missed my dad's company. I missed the bike rides. I missed \\
10 &  us were very sad. I missed the bike rides. I missed my dad. I missed my mom \\
15 &  us were very sad. My mom and I decided to go for a ride on our own. We \\
20 &  us were very sad. My mom and I decided to go for a ride on our own. We \\
30 &  us were very sad. My dad was the one who taught me to ride a bike. He was \\
50 &  us were very sad. My dad was the one who taught me how to ride a bike. He \\
100 &  us missed the bike rides. I was the one who missed it most. I was the one who \\
200 &  us missed it so much. Now, two years later, my dad is back on his feet and \\
500 &  a sudden I found myself without a companion. I missed the rides and the fun. I missed my \\
1000 &  us missed those outings. It's not easy to go out and do such activities with just me and \\
2000 &  his biking spares were sold. I still miss biking. INCIDENT INCIDENT INCIDENT INCIDENT INCIDENT INCIDENT INCIDENT INCIDENT \\
3000 &  DAMAGEREENREENREENREENREENREENREENREENREENREENREENREENREENREEN
\\
4000 &  INCIDENT INCIDENT INCIDENT INCIDENT INCIDENT INCIDENT INCIDENT INCIDENT INCIDENT INCIDENT INCIDENT INCIDENT INCIDENT INCIDENT INCIDENT INCIDENT INCIDENT INCIDENT INCIDENT INCIDENT \\
5000 &  favorayo tob disg favor JsonRequest/DkIFn favor Clay ped favor Clay ped Clay ped Clay ped Clay ped \\
\bottomrule
\end{tabular}
\caption{Generation upon ablation for stable-12b. See Appendix \ref{sec:generation} for details.}
\label{tab:ablation_gen_stable-12b}
\end{table*}

\end{document}